\documentclass[letterpaper,11pt]{article}
\usepackage[utf8]{inputenc}
\usepackage[T1]{fontenc}

\usepackage[english]{babel}
\usepackage{authblk}
\usepackage[margin=1in]{geometry}

\usepackage{amsfonts}       
\usepackage{nicefrac}       
\usepackage{microtype}      
\usepackage{amsmath,bbm,bm,mathtools} 
\usepackage{float}
\usepackage{amssymb}
\usepackage{amsthm}
\usepackage{algorithm}
\usepackage{algorithmic}
\usepackage{booktabs}       
\usepackage{graphicx}
\usepackage{multirow}
\usepackage{subcaption}
\usepackage[colorlinks=true, allcolors=black]{hyperref}
\usepackage{xcolor}
\usepackage{comment}
\usepackage[shortlabels]{enumitem}
\usepackage[normalem]{ulem}

\usepackage{natbib}
\setcitestyle{authoryear,open={(},close={)}}

\theoremstyle{plain}

\def\eqref#1{equation~\ref{#1}}

\def\1{\bm{1}}

\DeclareMathAlphabet{\mathsfit}{\encodingdefault}{\sfdefault}{m}{sl}
\SetMathAlphabet{\mathsfit}{bold}{\encodingdefault}{\sfdefault}{bx}{n}

\definecolor{DarkGreen}{rgb}{0.1,0.5,0.1}
\definecolor{DarkRed}{rgb}{0.5,0.1,0.1}
\definecolor{DarkBlue}{rgb}{0.1,0.1,0.5}
\definecolor{Gray}{rgb}{0.2,0.2,0.2}

\makeatletter
\newcommand\footnoteref[1]{\protected@xdef\@thefnmark{\ref{#1}}\@footnotemark}
\makeatother

\title{\Large{{Towards causal effect estimation with learned instrument representations}}}

\author{\large{Frances~Dean$^{*\dagger\S}$~~~Jenna~Fields$^{*\dagger\S}$~~~Radhika~Bhalerao$^{\dagger\S}$~~~ Marie~Charpignon$^{\dagger}$~~~Ahmed~Alaa$^{\dagger\S}$}}

\affil{$^*$Equal contribution \\
$^\dagger$University of California, Berkeley\\
$^\S$University of California, San Francisco}

\usepackage{hyperref}
\hypersetup{
  colorlinks=true,
  linkcolor=blue,
  citecolor=blue,
  urlcolor=blue
}

\usepackage{url}

\usepackage{multirow}
\usepackage{makecell}
\usepackage{mdframed}

\usepackage{comment}
\usepackage[utf8]{inputenc} 
\usepackage[T1]{fontenc}    
\usepackage{hyperref}       
\usepackage{url}            
\usepackage{booktabs}       
\usepackage{amsfonts}       
\usepackage{nicefrac}       
\usepackage{microtype}      
\usepackage{xcolor}         
\usepackage{tikz}
\usepackage{tikz-cd}
\usetikzlibrary{positioning}
\usepackage{soul}
\usepackage{amssymb}
\usepackage{amsthm}
\usepackage{wrapfig}
\usepackage{amsmath}

\theoremstyle{plain}

\date{}

\begin{document}

\maketitle

\begin{abstract}
Instrumental variable (IV) methods mitigate bias from unobserved confounding in observational causal inference but rely on the availability of a valid instrument, which can often be difficult~or~infeasible to identify in practice. In this~paper,~we propose a representation learning approach that constructs {\it instrumental representations} from observed covariates, which {could} enable IV-based estimation even in the absence of an explicit instrument. Our model (ZNet) achieves this through an architecture that mirrors the structural causal model of IVs; it decomposes the ambient feature space into confounding and instrumental components, and is trained by enforcing {\it empirical conditions} corresponding to the defining properties of valid instruments (i.e., relevance and exclusion restriction).~ZNet is compatible with a wide range of downstream two-stage IV estimators of causal effects. Our experiments demonstrate that ZNet (i) can recover ground-truth instruments when they already exist in the ambient feature space and (ii) constructs candidate latent instruments in the embedding space when no explicit IVs are available. {However, instrument representations have inherent challenges which warrant caution and further research.} This work {explores} when ZNet {might} be used as a module for causal inference in general observational settings. 
\end{abstract}

\section{Introduction}
Instrumental variable (IV) regression is a commonly used approach to address unobserved confounding when estimating causal effects from observational data. An IV randomizes the treatment without having a direct effect on the outcome, enabling estimation of causal effects with two stage regression \citep{Imbens_Rubin_2015}. For example, in economics, geographical proximity to a college is used as an instrument for educational attainment in estimating the returns of schooling \citep{card1993using} and draft lotteries as instruments to study the effect of military service on long-term economic outcomes \citep{angrist1990lifetime}. In medicine, genetic variants have been used as IVs since they often highly correlate with risk factors but do not directly influence the outcomes associated with these factors \citep{davey2014mendelian}. While IV regression enables causal identification in observational data, these methods are not ubiquitous because~a~suitable instrument must be available and known to the analyst. This is seldom the case in many~applications.~For~instance,~in health research, genetic variants are rarely available in large datasets, and moreover, candidate genes may not be strong or valid IVs \citep{davies2015many, burgess2016combining}.   

\begin{figure}[t]
\centering
\includegraphics[width=2.8in]{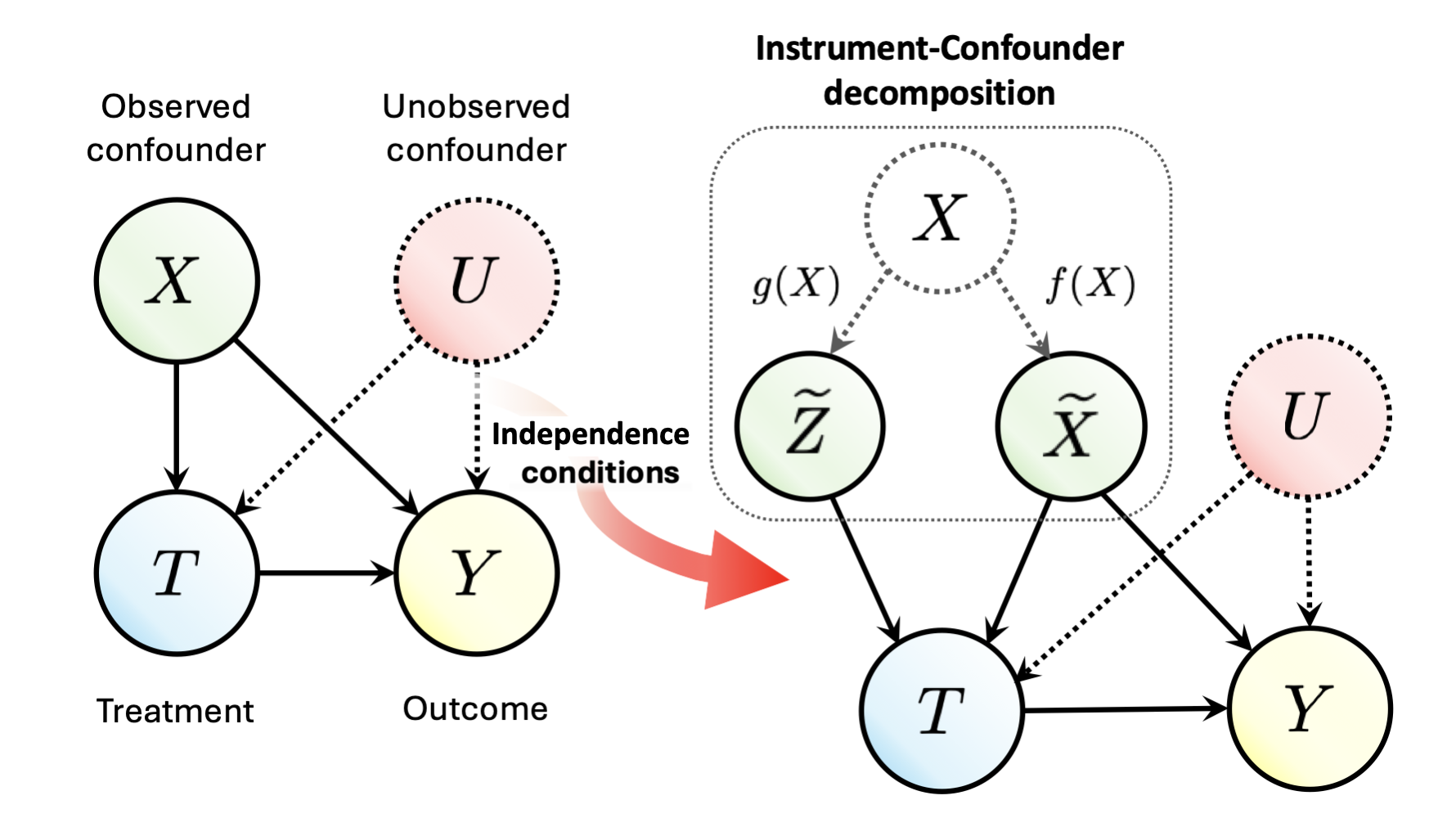} 
\caption{{\footnotesize {\bf Causal inference with learned instruments.} We learn a feature representation of the observed data \mbox{\scriptsize $X$} that decomposes them into a learned instrument \mbox{\scriptsize $\widetilde{Z} = g(X)$} and a residual confounder \mbox{\scriptsize $\widetilde{X} = f(X)$}. This decomposition is induced by encouraging conditions needed for the learned instrument \mbox{\scriptsize $\widetilde{Z}$} to be valid.}}
\vspace{.025in}
\label{fig:2}
\rule{\linewidth}{0.4pt}
\vspace{-.3in}
\end{figure}

The tradition of IV regression originated~in~settings~with~tabular data, where an instrument is expected to be~explicitly~observed within the feature space. In contrast, the increasing use of deep learning representations of high-dimensional data, such as text, images, and multimodal inputs, creates new opportunities to uncover instrumental variables that are implicit rather than directly observed. For~example,~medical images may encode subtle provider- or institution-specific patterns, while rich electronic health records may imitate genetic or phenotypic information through descriptions embedded in clinical notes. Representation learning methods could be used to extract such latent variables as instruments, thereby automating instrument selection and further extending IV methods to non-tabular settings where no~single~variable in the ambient feature space is a valid IV on its own.

In this paper, we develop a representation learning approach that constructs IVs from observed data~by~decomposing~the input feature space into confounding and instrumental components (Fig. \ref{fig:2}). Our model, which we call ZNet, is an encoder architecture that mirrors the {\it structural causal model} (SCM) of IVs and is regularized by enforcing {\it empirical conditions} that correspond to the defining properties of valid instruments (i.e., {relevance and exclusion restriction}).~The~learned~instrumental representation {could} then be used as input to standard two-stage IV estimators to recover causal effects. When a valid IV is present in the ambient feature space, ZNet recovers a representation that is strongly correlated with it; when no explicit instrument exists, ZNet learns a latent representation that could serve as a pragmatic instrument. The ZNet architecture is compatible with a broad class of downstream IV estimators, and through semi-synthetic experiments, we show that learning instrumental representations {can} reduce bias from unobserved confounding. Additionally, we illustrate the learned feature decomposition in real-world, high-dimensional settings using electrocardiogram data. {However, using IV representations rather than real IVs still has inherent challenges. IV representations in the absence of an underlying true instrumental variable are not guaranteed to be IVs, and any empirical success should be viewed cautiously.}

\section{Preliminaries}
\label{problem}
Let $Y \in \mathcal{Y} \subset \mathbb{R}$ denote a continuous outcome, $T \in \mathcal{T} \subset \mathbb{R}$ a treatment variable, and $X \in \mathcal{X} \subset \mathbb{R}^d$ a set of observed covariates associated with each unit. The treatment~$T$~has~a causal effect on the outcome $Y$, while $X$~may~influence~both $T$ and $Y$ ($X$ confound the relation of primary interest between $T$ and $Y$). In addition to the measured confounders $X$, there are unknown or {\it unobserved confounders} $U$, which induce spurious associations by simultaneously affecting both the treatment and outcome (Fig. \ref{fig:2}).~We~assume~that~the~outcome variable $Y$ is determined by the~following~SCM:
\begin{align}\label{Y_structure}
Y = \varphi(X,T) + e_Y(U), \,\,\,\,\,\,\,\, T = \psi(X) + e_T(U),
\end{align}
for unknown functions $\varphi : \mathcal{X} \times \mathcal{T} \rightarrow \mathcal{Y}$ and $\psi : \mathcal{X} \rightarrow \mathcal{T}$. Following \citep{hartford2017deep}, we assume that the unobserved confounders $U$ influence the outcome~$Y$~and~the treatment $T$ additively, via the error functions of U, which we henceforth denote $e_Y$ and $e_T$.~As~a~result,~the observational and interventional distributions generally differ, i.e., 
\begin{align}
\mathbb{E}[Y|X, T] &= \varphi(X,T) + \mathbb{E}[e_Y|X,T] \nonumber \\
                   &\neq \varphi(X,T) + \mathbb{E}[e_Y|X] = \mathbb{E}[Y|do(T), X]. \nonumber  
\end{align}
Thus, standard regression would lead to confounding bias.

{\bf De-confounding with instrumental variables (IVs).} A common method for removing confounding bias is to use IV regression. In the classical IV setting, we assume~access~to an additional variable $Z$ that is not~influenced~by~the~unobserved confounders $U$, affects the treatment $T$, and has no direct effect on the outcome $Y$ \citep{schooling_source, angrist1996}. Formally, given a set of observed~confounders~$X$, $Z$ is a valid IV if it satisfies the following conditions:

\vspace{-.275in}
\begin{align} \label{eq::assumptions}
\mbox{{\it Relevance:}}&\,\, Z \not \perp T\,|\,X, \nonumber \\  
\mbox{{\it Exclusion restriction:}}&\,\, Z \perp Y \,|\, (X, T, U), \nonumber \\ 
\mbox{{\it Unconfoundedness:}}&\,\, Z \perp U \,|\, X. 
\end{align}
Under these conditions, the instrument $Z$ can be used in a two-stage regression to estimate the effect of $T$ on $Y$. Under the additive model in (\ref{Y_structure}), \citet{hartford2017deep} uses~the~instrument $Z$ to set up an inverse problem by relating~the~counterfactual $\mathbb{E}[Y|do(T), X]$ to observable distributions:
\begin{align}\label{inverseproblem}
\mathbb{E}[Y|X, Z]&= \mathbb{E}[\varphi(X, T) + e_Y|X, Z] \nonumber\\&= \mathbb{E}[\varphi(X,T)|X,Z]+  \mathbb{E}[e_Y|X] \nonumber \\
        &=\mbox{$\int$}\, \mathbb{E}[Y|do(T), X]\, dF(T|X, Z).
\end{align}
Thus, with the instrument $Z$, we can estimate the counterfactual $\mathbb{E}[Y|do(T), X]$ by estimating the~two~observable functions $\mathbb{E}[Y|X, Z]$ and $F(T|X,Z)$. While this~inverse~problem is ill-posed, it provides a practical framework for estimating counterfactuals, and identification is possible under certain conditions \citep{newey2003instrumental}.\footnote{For example, in the linear case, Two-Stage Least Squares Regression (TSLS) allows for identification of causal effects.}~A two-stage regression fits a model $\widehat{F}(T|X,Z)$, and estimates $\mathbb{E}[Y|do(T), X]$ by replacing $F(T|X,Z)$ with $\widehat{F}(T|X,Z)$ in (\ref{inverseproblem}). With an estimate of $\mathbb{E}[Y|do(T), X]$, we can derive the conditional average treatment effects (CATE) and average treatment effects (ATE): 
\begin{align}
    \operatorname{CATE}(X) &= \mathbb{E}[Y|do(T)=1, X] - \mathbb{E}[Y|do(T)=0, X], \nonumber \\ 
    \operatorname{ATE} &= \mathbb{E}[Y|do(T)=1] - \mathbb{E}[Y|do(T)=0].
\end{align}
Equation (\ref{inverseproblem}) is notably more general than IV regression with linear models. We allow $\mathbb{E}[e_Y|X]\neq 0$, i.e., observed confounders can correlate with unobserved errors. In two stage least square regression (TSLS), confounders cannot be endogenous for unbiased estimates \citep{schooling_source}. 

\begin{figure*}[t]
\centering
\includegraphics[width=5.75in]{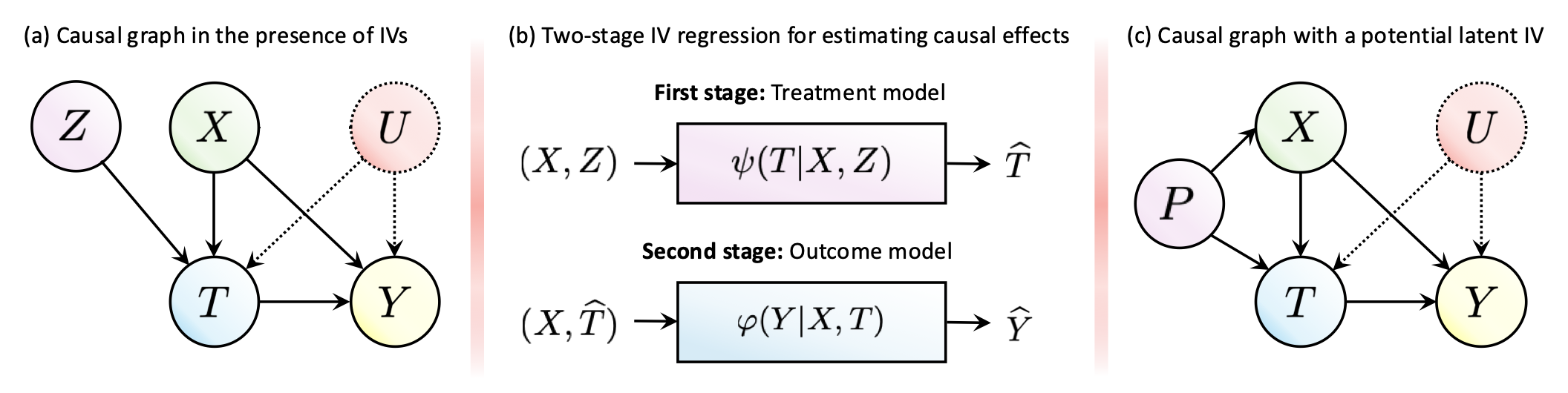}
\caption{\textbf{Illustration of the IV setting.} (a) Causal graph for the IV setting; (b) When an IV exists, causal effects can be estimated under unobserved confounding by first fitting a model $\psi$ to predict $T$ from $(X,Z)$, and then training a model $\varphi$ that predicts outcomes based on predicted treatments. (c) Illustration for the latent instrument example discussed in Section \ref{IC_decompose}. Here, $P$ is the provider identity.}
\label{fig:1}
\rule{\linewidth}{0.4pt}
\vspace{-.2in}
\end{figure*}

\section{Learning Instruments from Data}
\label{IC_decompose}
In the absence of a (known) valid~instrument~$Z$,~our~method {\it learns} an instrumental representation directly from the features $X$ using an observational dataset $\{(X_i, T_i, Y_i)\}_i$.

{\bf Instrument-confounder decompositions.} As illustrated in Fig. \ref{fig:2}, we seek a pair of functions $(f(\cdot), g(\cdot))$ that decompose the observed features $X$ into an observed {\it confounding} component $\widetilde{X}$ and an {\it instrumental} component $\widetilde{Z}$, i.e.,
\begin{align}\label{eq::decomposition} 
(\widetilde{X}, \widetilde{Z}) = (f(X), g(X)),    
\end{align}
such that $g(X)$ satisfies the three IV conditions in (\ref{eq::assumptions}), and the function $f(X)$ fits the additive SCM in (\ref{Y_structure}), i.e.,
\begin{align} 
Y = \varphi(f(X),T) + e_Y(U), \,\,\,\, T = \psi(f(X), g(X)) + e_T(U). \nonumber
\end{align}
Notice that this construction is broadly applicable and captures many real-world scenarios for how instrumental variations may be encoded in the feature space. First, consider the scenario where a subset of the observed variables can already serve as a valid instrument, i.e., $Z \subset X$.~In~this~setting, the decomposition trivially reduces to $g(X) = Z$ and $f(X) = X \setminus Z$. Second, consider scenarios in which~the~instrument is latent but can be inferred from proxy~signals~in the observed features $X$. For example, in~healthcare settings, providers are often quasi-randomly assigned and have different propensities to administer treatments based on their interpretation of patient information, while provider identity itself does not directly affect outcomes.~Although~provider identifiers may not be explicitly recorded, their influence shows in retrospective data through visit timing, laboratory and diagnostic tests ordered, and clinical notes (Fig. \ref{fig:1}(c)). A representation of provider influence can therefore be learned from $X$ via $g(\cdot)$ and used as an instrumental variable.

While the instrument–confounder decomposition is motivated by real-world scenarios, our model does not require an {\it a posteriori} interpretation of the learned instrument $\widetilde{Z}$. True satisfaction of the three IV conditions would allow for a pragmatic implementation of IV regression independent of interpretation, since our decomposition treats instruments as abstract mathematical constructs in a latent embedding space, even when no explicitly identifiable instrumental variable exists. 

{\bf Conditions on \boldsymbol{$f(.)$} and \boldsymbol{$g(.)$}.} A valid decomposition of the observed features $X$ into confounding and instrumental components must induce a factorization of the joint distribution $\mathbb{P}(g(X), f(X), U, T, Y)$ that is consistent with the SCM defined by (\ref{Y_structure}) and (\ref{eq::assumptions}). We train a network to construct functions through regularization using empirical constraints. In particular, the relevance condition suggests that the instrumental component $g(X)$ be predictive of the treatment, i.e., we train a network $\pi$ to be the best predictor of $T$ from $g(X)$. Exclusion restriction requires that all direct effects of $X$ on the outcome $Y$ be mediated through the confounding component $\widetilde{X} = f(X)$; i.e., $Y$ is well predicted by $f(X), T$, and $g(X)$ does not enter the equation except through $T$. To encourage this, we build a network $\phi$ predicting $Y$ from $f(X), T$ and make $g(X)$ independent from $f(X)$ and the residual $\widetilde{\varepsilon}_{Y} := Y-\mathbb{E}[Y | X, T]$. As we discuss later in detail in Section \ref{sec::znet}, we encourage these relationships through the architecture and loss function of our proposed model. 

\textit{Remark}. We remark that these conditions are not equivalently the IV conditions. First and foremost because we encourage these as empirical constraints, our work is not theoretical but \textit{empirical} in nature. {Moreover, as we do not redefine the relationships of $g(X), f(X)$ to $Y, T$ as in Equation \ref{Y_structure}, we do not completely force the IV conditions on $g(X), f(X)$. In particular, the conditions are not necessarily satisfied on $f(X)$ and $g(X)$ as random variables.} For example, $g(X)$ does not automatically reflect a variable that affects $Y$ {through} $T$ as it may be correlated or predictive only of another part of $T$.  

The learned instrument $\widetilde{Z}$ will be uncorrelated with the error $e_Y$ and satisfy unconfoundedness as long as $X$ is unconfounded by $U$ since it is derived directly from $X$. The assumption that $X$ is not influenced by $U$ is standard in other attempts at IV representations and allow for classical IV regression and straightforward IV generation \citep{yuan2022autoiv, li2024distribution, cheng2023learning, chou2024estimate}. To explore capabilities beyond this strong assumption, we include experiments with data that has confounding of $X$ by $U$. {To sum~up, for information denoted $I$, we desire~$f$ and $g$ to satisfy the following restrictions to learn an empirical approximation of an IV:}

\textbf{Condition 1}: $\mbox{I}(g(X); \tilde{\varepsilon}_Y)=0$.

\vspace{-.025in}
\textbf{Condition 2}: $\mbox{I}(g(X); f(X))=0$. 

\vspace{-.025in}
\textbf{Condition 3}: $\mbox{I}(Y; (f(X), T))>0$.

\vspace{-.025in}
\textbf{Condition 4}: $\mbox{I}(g(X); T)>0$. 

\section{Related Work}\label{related}
Prior work on learning IVs has largely focused on the problem of IV selection from a set of observed candidates. For instance, ModeIV chooses instruments by looking at clusters of treatment effects based on weighting the observed variables used as instruments \citep{hartford2021valid}. DIV-VAE uses a variational autoencoder (VAE) approach to disentangle an instrumental variable under the assumption that a surrogate instrument exists in the data \citep{cheng2024disentangled}. IV-Tetrad \citep{silva2017learning} builds strong instruments requiring at least two observed IV candidates. Several methods for refining IV candidates and estimating causal effects from these candidates can fall under this category as well, including the sisVIVE \citep{kang2016instrumental}, TEDVAE \citep{zhang2021treatment} and Ivy models \citep{kuang2020_Ivy} among various others \citep{davies2015many, burgess2016combining}.

Methods to learn IVs were also proposed in prior work, including GIV \citep{wu2023learning}, AutoIV \citep{yuan2022autoiv}, VIV \citep{li2024distribution}, DVAE-CIV \citep{cheng2023learning}, and GDIV \citep{chou2024estimate}. GIV generates a categorical IV with unsupervised expectation-maximization~that~groups data according to underlying distributional differences assumed to arise from the aggregation of data from multiple sources. This approach uses environment as an IV in data coming multiple sources \citep{schweisthal24ICML}. The other methods learn variational distributions. The AutoIV method uses a mutual information (MI) based loss to generate an abstract IV from observed data by learning variational distributions \citep{yuan2022autoiv}. VIV, DVAE-CIV, and GDIV use VAEs to learn independent latent variables that serve as $Z, U, C$ from the observed data $Y,T,X$, sometimes including an additional adjustment variable $A$ derived from $Y,X$. VAEs have shown great success in probabilistic modeling in general but {also} lack theory to guarantee learning the true causal model and satisfaction of IV conditions.

\section{The ZNet Model}
\label{sec::znet}

\subsection{ZNet Architecture}
\begin{figure}[t]
    \centering
 \includegraphics[width=3in]{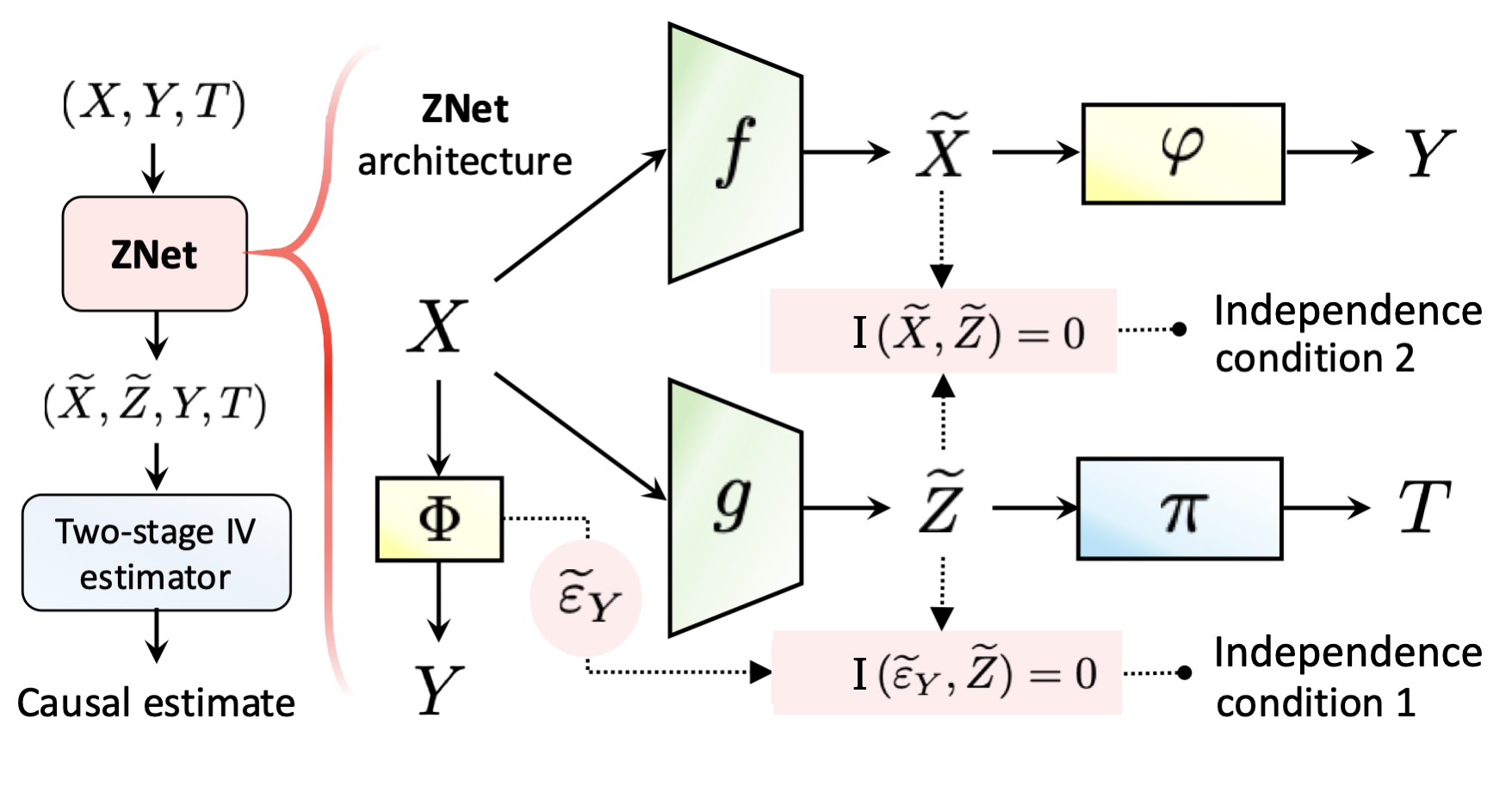} 
    \caption{{\bf Overview of the components of the ZNet Architecture.}}
    \label{fig:ZNetPCMI}
    \vspace{.05in}
    \rule{\linewidth}{0.4pt}
    \vspace{-.4in}
\end{figure}

The ZNet architecture encodes an inductive prior consistent with the SCM of the IV setup in (\ref{Y_structure}). The ZNet model comprises two encoders, $f$ and $g$, which learn representations of the confounders $\widetilde{X}$ and the instrument $\widetilde{Z}$, respectively, along with three feedforward neural networks, $\Phi, \varphi$ and $\pi$, used to estimate $\mathbb{E}[Y|X,T], \mathbb{E}[Y|\widetilde{X},T]$ and $\mathbb{E}[T|\widetilde{Z}]$. An overview of all components of the ZNet model is shown in Fig. \ref{fig:ZNetPCMI}. After training, ZNet produces~embeddings~$\{\widetilde{X}, \widetilde{Z}\}$ from the observed features $X$, which we then use to transform the observational dataset $\mathcal{D}=\{(X_i, T_i, Y_i)\}_i$~into~a~new dataset $\widetilde{\mathcal{D}}=\{(\widetilde{X}_i, \widetilde{Z}_i, T_i, Y_i)\}_i$ with the learned instruments $\{\widetilde{Z}_i\}_i$. As we show in the experiments section, ZNet is compatible with any downstream IV-based causal estimator. 

\subsection{ZNet Loss Function}
We first train the network $\Phi$ to obtain the regression residual $\tilde{\varepsilon}_Y := Y - \mathbb{E}[Y|X,T]$, and then train the ZNet model end-to-end by minimizing the predictive loss of the supervised learning models $\varphi$ and $\pi$, and imposing the conditions on the learned embeddings as constraints $(\widetilde{X}, \widetilde{Z})$ as discussed~in~Section~\ref{IC_decompose},~i.e.,
\begin{align}
\min_{f,g,\varphi,\pi} \mathcal{L}_Y(\varphi(f)) + \mathcal{L}_T(\pi(g))
\,\, \text{s.t.} \,
\begin{cases}
\operatorname{I}(g(X), \tilde{\varepsilon}_Y) = 0, \\
\operatorname{I}(f(X), g(X)) = 0. 
\end{cases} \nonumber
\end{align} 
Here, $\mathcal{L}_Y(\varphi(f))$ and $\mathcal{L}_T(\pi(g))$ are supervised~losses for the models $\varphi(f(X), T)$ and $\pi(g(X))$ against the labels $Y$ and $T$ (e.g., mean squared error and cross entropy losses). Note that conditions 3 and 4 in Section \ref{IC_decompose} are implicit in these losses. We train ZNet by replacing the hard constraints with soft penalty terms as follows:
\begin{align}
\min_{f,g,\varphi,\pi} \mathcal{L}_Y(\varphi(f)) + \mathcal{L}_T(\pi(g)) &+ \lambda_0 \operatorname{I}(g(X), \tilde{\varepsilon}_Y) \nonumber \\ 
&+ \lambda_1 \operatorname{I}(f(X), g(X)),\nonumber
\end{align} 
where $\lambda_0$ and $\lambda_1$ are tunable hyperparameters. Our training algorithm approximates the information terms $I$ above using Pearson correlation coefficients or mutual information, as described in the next section.

\subsection{ZNet Implementation and Training Details} 
The networks $\Phi, \varphi, \pi, f, g$ are implemented as feedforward networks, where the activation function can be chosen between ReLU or linear. Supervised learning losses for $\Phi, \varphi$ and $\pi$ are implemented as mean squared error (MSE), MSE, and binary cross entropy (BCE), respectively. 

{\bf Empirical approximation of independence constraints.}~The~information terms in the ZNet loss function are approximated through either the squared Pearson correlation or an approximation of the mutual information (MI) using kernel density estimation (KDE) with Gaussian kernels {since we consider small data experiments}; the selection of the approximation method is treated as a hyperparameter. We augment the standard supervised losses for $\varphi, \pi$ with the option of learning {networks \textit{f}, \textit{g}} through constraints on shared information as well. In this case, each network's loss is implemented as the square complement of Pearson correlation or negative MI. To regularize the distribution of the resulting IV~$\tilde{Z}$ and confounder $\tilde{X}$,~we~add distribution restrictions using a Kullback-Leibler (KL) divergence penalty on each dimension of $\tilde{X}$ and $\tilde{Z}$ with a standard normal distribution $\mathcal{N}(0,1)$, denoted as KL. We also constrain the average correlation across dimensions within $\tilde{X}$ and $\tilde{Z}$~to~encourage the features of the learned representations to be distinct, denoted by $\operatorname{C}$. The overall loss with an approximation of I as \mbox{\small $\widehat{\operatorname{I}}(.)$} is thus given as:
\vspace{-.08in}
\begin{align*}
 \mathcal{L}_{\mbox{ZNet}}& =  \alpha_1 \operatorname{MSE}(\varphi(f(X), T), Y)\\& + \alpha_2 \operatorname{BCE}(\pi(g(X)), T)\\
   &+ \alpha_3 (-\widehat{\operatorname{I}}(f(X), Y)) \\&+ \alpha_4(- \widehat{\operatorname{I}}(g(X), T))\\
   &+ \alpha_5 \widehat{\operatorname{I}}(g(X), \tilde{\varepsilon}_Y) + \alpha_6 \widehat{\operatorname{I}}(g(X), f(X)) \\
   &+ \alpha_7\operatorname{KL}(g(X), \mathcal{N}) + \alpha_8\operatorname{KL}(f(X), \mathcal{N}) \\&+ \alpha_9 \operatorname{C}(g(X)) + \alpha_{10} \operatorname{C}(f(X))\\& + \alpha_{11} \operatorname{MSE}(\Phi(X,T), Y).
\end{align*}

{\bf Training steps.} ZNet training occurs in three stages. First, the network $\Phi$ is separately trained to estimate $\tilde{\varepsilon}_Y$, i.e. only $\alpha_{11}$ is non-zero. Weights of $\Phi$ are then frozen. Next the full network is pretrained by fitting $\varphi, \pi$ in a supervised setting without the constraints ($\alpha_3=\alpha_4=\alpha_5=\alpha_6=\alpha_7=\alpha_8=\alpha_9=\alpha_{10} = \alpha_{11}=0$). This serves to encourage starting representations of $\tilde{X}$ and $\tilde{Z}$ to be highly correlated with $Y$ and $T$, respectively. Finally, the full ZNet model is finetuned end-to-end across all the loss terms above (with $\alpha_{11}=0$). 

{\bf Hyperparameter tuning.} All hyperparameters, including all loss term weights and whether the independence constraints are approximated via Pearson correlations or MI, were tuned using Bayesian optimization implemented in Botorch \citep{balandat2020botorch}. We perform optimization corresponding to the two stages of IV regression. For each IV generation, we maximize the instrument's relevance F-Statistic and minimize the correlation between learned $\tilde{X}$ and $\tilde{Z}$ using Botorch's native adaptation of the Noisy Expected Improvement acquisition function for multi-objective optimization. We then choose the parameter set from the Pareto front with the highest F-Statistic. We tune the treatment effect estimators to simultaneously minimize the MSE of the model's ATE against a nearest-neighbors (NN) ATE and the MSE of estimated factual $Y$, again with the Noisy Expected Improvement acquisition function. The parameter~set~is~selected from the Pareto front by the smallest MSE of NN. 

\subsection{Causal Effect Estimation}
Recall that ZNet creates a dataset with learned confounders and instruments, and is therefore compatible with a wide range of downstream IV-based estimators of causal effects. We wrap our ZNet model around three downstream estimators of treatment effects to demonstrate the utility of ZNet in causal inference: TSLS, DFIV and DeepIV. Each method takes as input the true treatment term $T$ and our~learned~representations $\tilde{X}$ and $\tilde{Z}$. TSLS is the classical IV estimator. It assumes linear structural equations and independence of $U$ and $X$, thus we omit when not applicable \citep{Imbens_Rubin_2015, schooling_source}. DeepIV \citep{hartford2017deep} generalizes TSLS by allowing the model at each stage to be parameterized by a neural network and $X\not \perp U$. The DFIV estimator allows basis functions at each stage to be parametrized by neural networks \citep{xu2020learning}. 

\section{Experiments}\label{eval}
{\bf Datasets.} We conduct a comprehensive and extensive  evaluation experiments of ZNet using the semi-synthetic \textbf{IHDP} dataset, a widely used benchmark for causal inference \citep{hill2011bayesian}. IHDP is based on an experiment that~studied~the~effect of home visits during infancy on cognitive test scores~of premature infants, with 985 individuals and 25 covariates. We design multiple experimental setups using IHDP that correspond to different ways in which IVs may explicitly or implicitly exists. Each setup is define by three sets of features, $X^{\to T}$, $X^{\to Y}$, and $X^{\leftarrow U}$, where we each $X^{\to I}$ is the subset of covariates $X$ which have causal influence on $I$ in the arrow's direction. We create four experimental setups based on their inclusion of an instrument: 

{\footnotesize
\vspace{.1in}
\begin{mdframed}[linewidth=0.8pt]
\vspace{.025in}
\textbf{$\bullet$\,\, Disjoint Candidate:}\, $\exists \; X^{\to T} \;\text{s.t.}\;$ 
    \[
    X^{\to T} \cap X^{\to Y} = \varnothing,
    X^{\to T} \cap X^{\leftarrow U} = \varnothing.
    \]
\textbf{$\bullet$\,\, Mixed Candidate:}\, $\exists \; \widetilde{X}^{\to T} \subset X^{\to T} \;\text{s.t.}\;$
    \[
    \widetilde{X}^{\to T} \cap X^{\to Y} = \varnothing,
    \quad
    \widetilde{X}^{\to T} \cap X^{\leftarrow U} = \varnothing.
    \]
\textbf{$\bullet$\,\, Latent Categorical Instrument:}\, $    \exists \; Z, f \;\text{s.t.}\;$
    \[
    f(X^{\to T}) = Z \in \mathbb{N}^{+}.
    \]
\textbf{$\bullet$\,\, No Candidate:} $\nexists \; \widetilde{X}^{\to T} \subseteq X^{\to T} \;\text{s.t.}\;$
    \[
    \widetilde{X}^{\to T} \cap X^{\to Y} = \varnothing,
    \quad
    \widetilde{X}^{\to T} \cap X^{\leftarrow U} = \varnothing.
    \]
\end{mdframed}}

\begin{figure*}[t!]
    \centering 
\includegraphics[width=.95\linewidth]{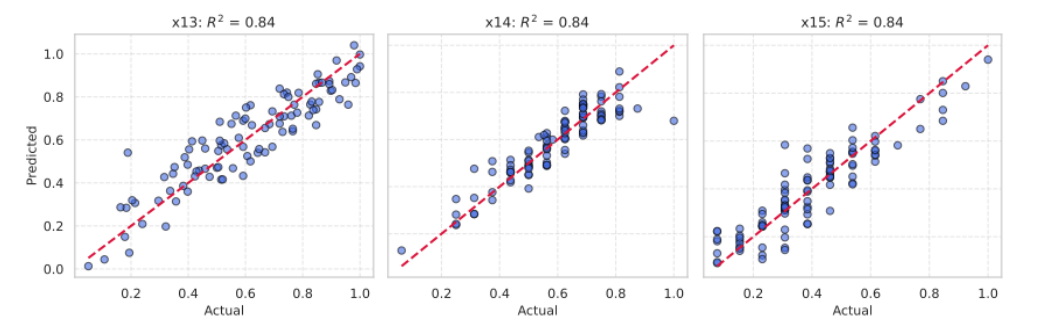}
    \caption{\textbf{Regression plots for learned and true instruments.} Learned instruments are correlated to true ones in the setup with~true~instrument candidate. Here, $X_{13}, X_{14}, X_{15}$ are true IVs ($x$-axis) plotted against the corresponding dimensions of the learned $\widetilde{Z}$ ($y$-axis).} 
    \label{fig:recovery_linear}
    \rule{\linewidth}{0.4pt}
    \vspace{-.24in} 
\end{figure*}

Directed acyclic graphs for all settings above are included in the Appendix. For each setting, we consider~scenarios~where $X^{\leftarrow U} \neq \varnothing$ and $X^{\leftarrow U}= \varnothing$. We also consider data where $U=\varnothing$ (i.e. no unobserved confounding). After~fixing~a~covariate set, we choose functions $\phi, \psi, e_Y, e_T$ and generate the variables $Y,T$ similar to \citep{wu2023learning} by writing
\begin{align*}
    Y & =\phi(X_Y, T) + e_Y(U) +\varepsilon_Y \mbox{ for } \varepsilon_Y\sim \mathcal{N}(0,.1), \\
    T &\sim \mbox{Bernoulli}(P) \mbox{ for } P =  \psi(X_T) + e_T(U) + \varepsilon_T, \nonumber
\end{align*}
for $\varepsilon_T \sim \mathcal{N}(0,.1)$. We focus on binary treatments, though ZNet is easily adapted for continuous settings. We consider linear and non-linear version of $\phi, \psi$ for each dataset. For each scenario, we randomly generate 10 datasets with varying functions $\phi, e_Y, \varepsilon_Y, \psi, e_T, \varepsilon_T$. Across dataset and confounding settings, this means we evaluate ZNet on a comprehensive collection of 180 datasets. Data~are~always~split into 60\% for training, 20\% for tuning, and 20\% for testing. All experiments report results on the test proportion.

{\bf Metrics and baselines.} All model performances were evaluated using bootstrapping (N=50). ATE errors are evaluated in terms of absolute proportions: $|\mbox{Estimate} - \mbox{True}|/|\mbox{True}|$. Error on CATE estimation is presented as average square error in estimated individual level effect (PEHE). In addition to prior methods discussed in Section \ref{related}, we compare ZNet-learned instruments to using the true instrument \textit{TrueIV}, if it exists, and to TARNet \citep{shalit2017estimatingindividualtreatmenteffect}, a CATE estimation model used in the absence of unobserved confounding.

\begin{table}[h!]
\centering
\caption{Evaluating the average instrumental properties of ZNet learned representations}
{\footnotesize
\begin{tabular}{ccc}
\toprule
\textbf{Metric} & \textbf{ZNet} & \textbf{TrueIV} \\
& (Train / Val / Test) & (Train / Val / Test) \\
\midrule
F-Stat($\tilde{Z}$,$T$) & 140.25 / 14.82 / 8.70 & 57.36 / 19.35 / 8.18 \\Corr($\tilde{Z}$,$\tilde{X}$) & 0.15 / 0.16 / 0.19 & 0.09 / 0.10 / 0.16 \\Corr($\tilde{Z}$,$U$) & 0.13 / 0.13 / 0.15 & 0.03 / 0.06 / 0.06 \\Corr($\tilde{Z}$, $\tilde{\varepsilon}_Y$) & 0.02 / 0.06 / 0.09 & 0.04 / 0.05 / 0.08 \\
\midrule
\bottomrule
\end{tabular}}
\label{tab:iv_props}
\vspace{-.15in}
\end{table}

\subsection{ZNet Instrument Representations}
\begin{figure*}[t!]
        \centering
        \includegraphics[width=.9\linewidth]{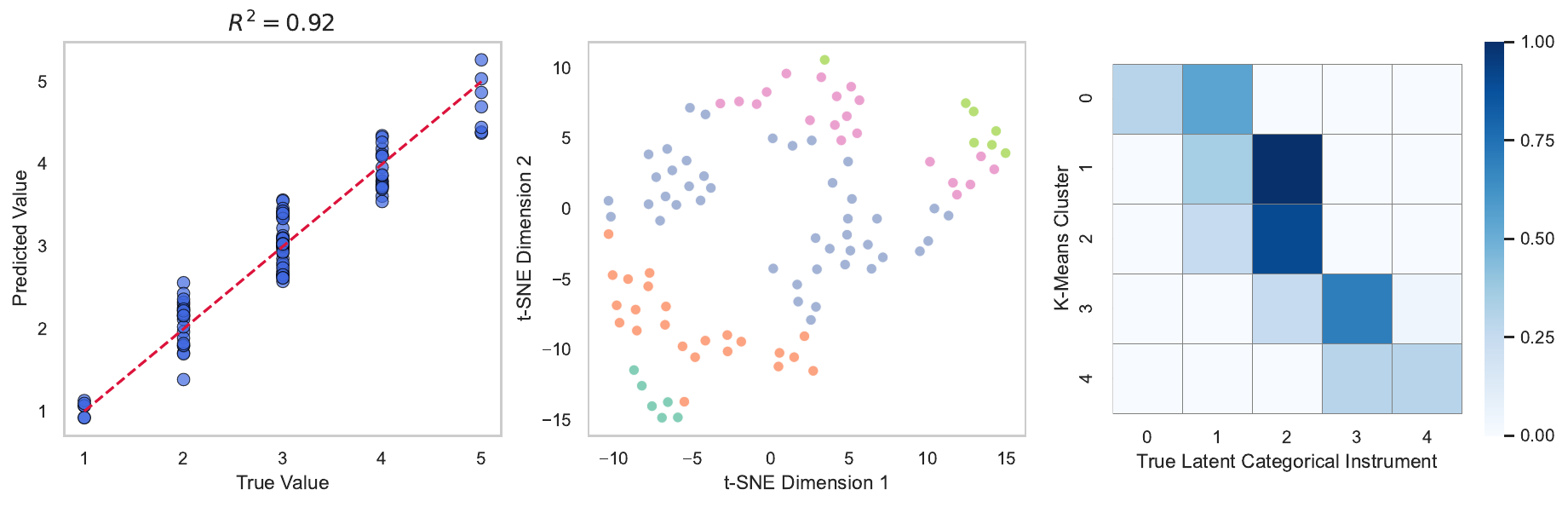}
    \caption{\textbf{Learned instrument representations are correlated to existing instruments in latent categorical instrument dataset.} Left most plot are regression predictions of the true categorical IV using learned $\widetilde{Z}$. Middle plot is a t-SNE visualization of $\widetilde{Z}$ colored by the true IV. Right most is a normalized confusion matrix of K-means clusters of $\widetilde{Z}$ compared to the true instruments.}
    \label{fig:giv}
    \vspace{.025in}
    \rule{\linewidth}{0.4pt}
    \vspace{-.25in}
\end{figure*}

As shown in Fig. \ref{fig:recovery_linear}, ZNet successfully recovers~existing~instruments. In the \textbf{Linear Mixed Candidate} setting, there are three variables $X_{13}, X_{14}, X_{15} \in X^{\to T}$ which are explicit instruments. In one selected sample dataset,~ZNet~generates a 10-dimensional variable $Z$ which is correlated with and linearly predicts each of $X_{13}, X_{14}, X_{15}$. ZNet also recovers latent instruments. Consider an example \textbf{Linear Categorical Instrument} dataset. The true instrument groups the observed data into 5 clusters. ZNet approximately recovers these clusters by regression, using t-SNE dimensionality reduction, and K-Means with cluster relabeling in Fig. \ref{fig:giv}.

\begin{table}
\vspace{.075in}
\centering
    \caption{Mean absolute ATE proportional error (SE) across all experimental settings with the DeepIV and DF-IV estimators.}
    {\footnotesize
    \begin{tabular}{llll}
        \toprule
        \textbf{Downstream} & \textbf{Model} & \textbf{Datasets} & \textbf{$\mid$ATE$\mid$ Error} \\
        \midrule
             & TARNet & 180 & 0.442 (0.03) \\
        \midrule
        \multirow{5}{*}{\makecell[l]{DeepIV}} 
 & TrueIV & 120 & 0.406 (0.03) \\
        \cmidrule(lr){2-4}
         & ZNet & \multirow{4}{*}{\makecell[l]{180}} & \textbf{0.412 (0.03)} \\
         & AutoIV &  & 0.440 (0.03) \\
         & VIV &  & 0.438 (0.04) \\
         & GIV &  & 0.429 (0.03) \\
        \midrule
        \multirow{5}{*}{\makecell[l]{DF IV}} 
 & TrueIV & 120 & 0.604 (0.09) \\
        \cmidrule(lr){2-4}
         & ZNet & \multirow{4}{*}{\makecell[l]{180}} & \textbf{0.544 (0.04)} \\
         & AutoIV &  & 0.705 (0.06) \\
         & VIV &  & 0.652 (0.11) \\
         & GIV &  & 0.798 (0.07) \\
        \midrule
        \bottomrule
    \end{tabular}}
    \label{tab:mean_ate_summary}
    \vspace{-.05in}
\end{table}

Independent of the existence of an instrument in the observed data, the ZNet model generates a representation that is correlated with $T$, independent of the confounder representation $\widetilde{X}$, independent of the error in predicting $Y$, and unconfounded by $U$ (as long as $U$ is not upstream of $X$). We evaluate the suitability of these instrument representations empirically. We observe F-Statistics that indicate the predictive power of generated representations $\widetilde{Z}$ for $T$ and low correlation across prohibited relationships between $\widetilde{Z}$ and confounders. ZNet satisfaction of empirical IV conditions is strong on average over all 180 datasets compared to the average over existing dataset IVs (Table \ref{tab:iv_props}). We additionally compare satisfaction by dataset scenarios in Appendix Tables \ref{tab:appendix-linear-instrument-strength}, \ref{tab:appendix-nonlinear-instrument-strength}. 

\subsection{Causal Effect Estimation with ZNet}
We used the ZNet learned instrument representations to recover the causal effects using a second stage regression with DeepIV or DFIV. We assess performance~across~the all experimental settings. We highlight performance across three comparisons: (1) all 180 datasets, (2) the 80 datasets with confounding influencing $X$, i.e. $U \to X$ datasets, and (3) the 40 confounded datasets with no instrument candidate. 

The first comparison shows that the performance of ZNet in estimating ATE outperforms baselines across the different experimental settings (Table \ref{tab:mean_ate_summary}). Notably, ZNet is the only method that {modestly} outperforms TARNet, which ignores unobserved confounding altogether. Additionally, ZNet performs best in the setting where the unobserved confounding influences the observed data, i.e. $U \to X$ datasets (Table \ref{tab:u_to_x_mean_ate_summary}). 

\begin{table}[h!]
\centering
    \caption{Mean absolute ATE proportional error (SE) across $U \to X$ datasets with the DeepIV and DF-IV estimators.}
    {\footnotesize
    \begin{tabular}{llll}
        \toprule
        \textbf{Downstream} & \textbf{Model} & \textbf{Datasets} & \textbf{$\mid$ATE$\mid$ Error} \\
        \midrule
             & TARNet & 80 & 0.666 (0.06) \\
        \midrule
        \multirow{5}{*}{\makecell[l]{DeepIV}} 
 & TrueIV & 60 & 0.534 (0.05) \\
        \cmidrule(lr){2-4}
         & ZNet & \multirow{4}{*}{\makecell[l]{80}} & \textbf{0.620 (0.06)} \\
         & AutoIV &  & 0.677 (0.06) \\
         & VIV &  & 0.702 (0.07) \\
         & GIV &  & 0.672 (0.06) \\
        \midrule
        \multirow{5}{*}{\makecell[l]{DF IV}} 
 & TrueIV & 60 & 0.736 (0.17) \\
        \cmidrule(lr){2-4}
         & ZNet & \multirow{4}{*}{\makecell[l]{80}} & \textbf{0.729 (0.08)} \\
         & AutoIV &  & 0.870 (0.10) \\
         & VIV &  & 0.906 (0.15) \\
         & GIV &  & 0.860 (0.09) \\
        \midrule
        \bottomrule
    \end{tabular}}
    \label{tab:u_to_x_mean_ate_summary}
    \vspace{-.1in}
\end{table}
The performance comparison in Table \ref{tab:no_cand_mean_ate_summary} focuses on the most common setting where explicit instrument candidates are not available. ZNet once again outperforms existing IV generation methods and TARNet. Performance on causal effect estimation was worse using DF IV than DeepIV consistently regardless of data generation method. We also include additional results stratified by the data generation settings which show where IV estimation is easier or more challenging and results of analyses with TSLS in the appendix (See Appendix Tables \ref{tab:appendix-ate-linear-results}, \ref{tab:appendix-ate-nonlinear-results}). We report~results on the PEHE in CATE estimation, which similarly varies by data generation settings, in the appendix as well (Appendix Tables \ref{tab:appendix-pehe-linear}, \ref{tab:appendix-pehe-nonlinear}).

\begin{table}[h!]
 \centering
    \caption{Mean absolute ATE proportional error (SE) across confounded settings with no true instrument candidate.}
    {\footnotesize
    \begin{tabular}{llll}
        \toprule
        \textbf{Downstream} & \textbf{Model} & \textbf{Datasets} & \textbf{$\mid$ATE$\mid$ Error} \\
        \midrule
             & TARNet & 40 & 0.619 (0.10) \\
        \midrule
        \multirow{4}{*}{\makecell[l]{DeepIV}} 
 & ZNet & \multirow{4}{*}{\makecell[l]{40}} & \textbf{0.566 (0.09)} \\
         & AutoIV &  & 0.667 (0.10) \\
         & VIV &  & 0.654 (0.11) \\
         & GIV &  & 0.704 (0.11) \\
        \midrule
        \multirow{4}{*}{\makecell[l]{DF IV}} 
 & ZNet & \multirow{4}{*}{\makecell[l]{40}} & \textbf{0.773 (0.14)} \\
         & AutoIV &  & 0.837 (0.14) \\
         & VIV &  & 0.986 (0.28) \\
         & GIV &  & 1.077 (0.17) \\
        \midrule
        \bottomrule
    \end{tabular}}
    \label{tab:no_cand_mean_ate_summary}
    \vspace{-.05in}
\end{table}

\subsection{Application of ZNet to Unstructured Data}
ZNet in the unstructured data setting is particularly powerful because high dimensional data can contain additional information recoverable as an instrument more readily. We believe this is especially impactful in the setting of unstructured health data. We demonstrate the ability of ZNet~to~recover candidate instruments in such settings using a semi-synthetic electrocardiogram (ECG) dataset. Data from 35,463 ECGs from a public Physionet dataset serve as $X$, age and sex serve as hidden confounders $U$, and variables $T$ and $Y$ are sampled, analogous to the process for IHDP data, in this case as a random linear combination of eight interpretable, derived features of the ECG, i.e. heart rate, QT interval, QRS amplitude, etc. \cite{ecg_data1, ecg_data2, physionet}. As with most unstructured health data, no true instrument exists. Despite that, ZNet still recovers an instrument representation and more accurately predicts treatment effects than ordinary least squares regression using the tabular features (Table \ref{tab:ecg_data}). Here, we use ZNet with Deep IV since it is consistently more accurate than DF-IV.

\begin{table}[h!]
    \caption{Mean absolute ATE proportional error (SE) across bootstraps in the ECG dataset.}

    \centering
    \begin{tabular}{lll}
        \toprule
        \textbf{Downstream} & \textbf{Model} &  \textbf{$\mid$ATE$\mid$ Error}\\
        \midrule
        DeepIV & ZNet & 0.126 (0.000647)  \\
        & OLS & 0.665 (0.00263)  \\
        \midrule
        \bottomrule
    \end{tabular}
    \label{tab:ecg_data}
    \vspace{-.1in}
\end{table}

\begin{figure}[h!]
\centering
\includegraphics[width=0.76\linewidth]{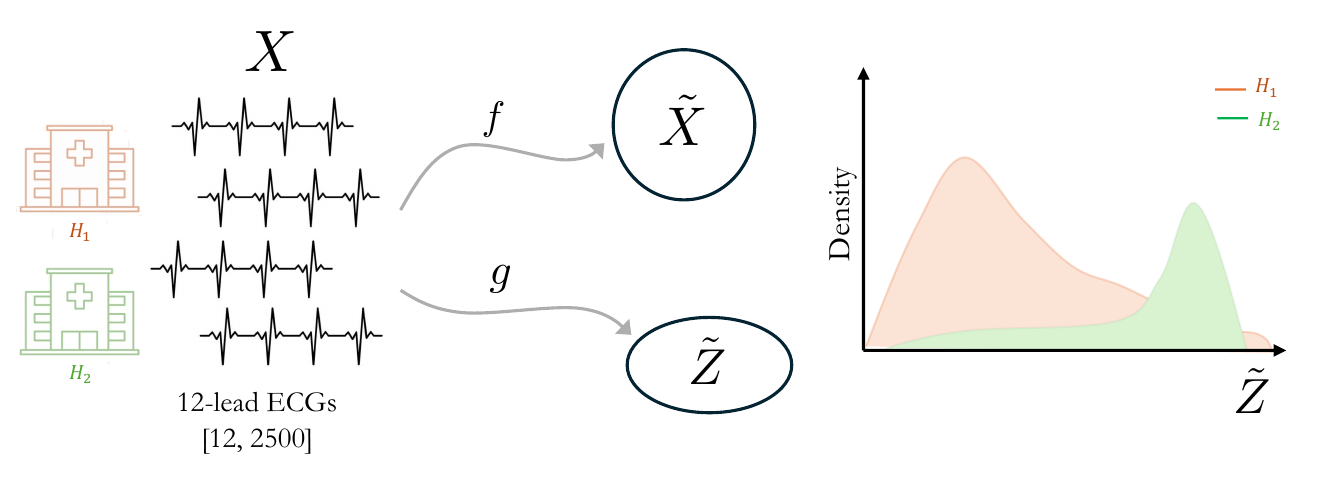}
    \caption{\textbf{Visualization of ZNet with unstructured health data.}}
       
    \label{ecg_figure}
   
    \rule{\linewidth}{0.4pt}
    \vspace{-.3in}
\end{figure}

These ECG data come from two hospitals in Zhejiang, China. In true observational data, both treatments received and images taken can vary by care setting due to purchased equipment or physician propensity despite being random with respect to many confounders. Moreover, the outcomes associated to treatments are typically uncorrelated with hospital setting directly. While hospital setting is not available in the ECG data and its instrumental value hidden to~an~analyst, with ZNet, we extract a representation that is encouraged to satisfy instrument properties potentially leveraging this latent information as illustrated pictorially in Fig. \ref{ecg_figure}.~We~evaluate~the instrument selected by ZNet by its satisfaction of IV properties empirically as correlations in Table \ref{tab:ecg_instrument_eval}. 

\begin{table}[h!]
    \caption{Evaluating the IV conditions of the ECG based instrument.}
    \centering
    {\footnotesize
    \begin{tabular}{cc}
        \toprule
        \textbf{Metric} &  \textbf{ZNet ECG IV} \\
        & (Train / Val / Test) \\
        \midrule
        F-Stat$(\tilde{Z}, T)$ & 261.26 / 59.69 / 22.57 \\
        $\mbox{Corr}(\tilde{Z}, \tilde{X})$ & 0.0069 / 0.0064 /  0.0076\\  
         $\mbox{Corr}(\tilde{Z}, U_1)$ & 0.25 / 0.25 / 0.244\\
         $\mbox{Corr}(\tilde{Z}, U_2)$ &  -0.15 / -0.16 / -0.15 \\ 
        $\mbox{Corr}(\tilde{Z}, \tilde{\varepsilon}_Y)$ & 0.0022 / 0.0048 / 0.015\\ 
        \midrule
        \bottomrule
    \end{tabular}}
    \label{tab:ecg_instrument_eval}
    \vspace{-.1in}
\end{table}

\section{Discussion}

{We consider this work an exploration of the potential of IV representations for causal inference.} ZNet {can} enable IV regression without domain knowledge of pre-existing IVs. This automation of IV generation from observed data is exciting, as it could enable the widespread use of IV regression for causal effect estimation. To this end, we contribute a fairly comprehensive evaluation of IV generation for causal inference, which demonstrates utility. Regardless of the existence of a candidate or a latent instrument or of unobserved confounding, ZNet {does} exceed the performance of TARNet and of probabilistic IV generation methods. This performance is enabled by the recovery of instrument representations, {which, while not guaranteed to be valid}, still show some empirical benefit in downstream inference. In the case of existing instruments among the observed data, we see that recovered instruments are highly correlated with these variables. Without candidates, ZNet representations are still relevant and show minimal correlation with hidden confounders and directly with the outcome, {but we cannot guarantee these representations are valid instruments}. 

The data generation process of any real world dataset is untestable. We can never confirm a lack of unobserved confounding, even when evaluating a candidate instrument. ZNet's strong performance across settings suggests that it can in some instances serve to mitigate unobserved confounding in general observational
settings, regardless of whether the untestable assumption of unconfoundedness is satisfied. Our evaluations show ZNet to be particularly well suited compared to other IV generation methods when unobserved confounders influence the observed data and where instrument candidates do not exist. These are the most challenging and general settings. ZNet does not, however, guarantee empirically in all cases that instruments are valid, and our work does not prove that ZNet representations are IVs, which greatly limits the applicability of the approach.

Further exploring inference with ZNet is especially relevant in the context of high dimensional unstructured data. Unstructured data may contain latent or abstract instruments more frequently, as high-dimensional feature spaces often contain rich information that our approach could learn to extract as instruments. High dimensional tabular and image data are becoming ubiquitous in the predictive models used across industries. Interest grows in embedding causal relationships into these models so that insights from their features are actionable. ZNet could automatically mitigate unobserved confounding broadening the ability to pursue causal machine learning in predictive settings if a valid instrument is detected. We illustrate ZNet's treatment effect estimation capability with ECGs, but applications for ZNet extend far beyond this modality and beyond healthcare. 

Solutions to the ZNet loss minimization problem do not give a representation that is guaranteed to be a valid instrument, a challenge reminiscent of those with representation learning in causal inference more generally \citep{locatello2019challenging}. We suspect future research pursuing guarantees on IV representations will be more limited when underlying instruments do not exist to be recovered. IV estimation is also limited in general by a lack of theoretical guarantees of identifiably without conditions on the SCM. This theoretically limits not only our approach but all deep learning based IV estimation. Moreover, we emphasize that our IV representation method does not and cannot mitigate unobserved confounding in all settings. Still, our experiments and opportunities in high dimensional data suggest the value in the increased research into the utility of these methods with caution in the absence of domain knowledge. 

\section*{Code}

All code for running ZNet is available at \url{https://github.com/AlaaLab/ZNet.git}.

\bibliography{biblio}
\bibliographystyle{plainnat}

\appendix
\newpage

\section{Additional Figures}

\begin{table}[h!]
\footnotesize
    \centering
    \caption{ATE Results on Synthetic Linear Datasets}
    \begin{tabular}{lccclcc}
        \toprule
        \textbf{Dataset} & \textbf{Diff Means} & \textbf{TARNet} & \textbf{TSLS} & \textbf{IV Method} & \textbf{DeepIV} & \textbf{DF IV} \\
        \midrule
        \multirow{5}{*}{\makecell[l]{Linear Disjoint}} 
            & 0.318 & 0.140 & 0.210 & TrueIV & 0.112 & \textbf{0.175} \\
            &  &  &  & ZNet & 0.110 & 0.229 \\
            &  &  &  & AutoIV & \textit{0.090} & 0.329 \\
            &  &  &  & VIV & \textbf{0.088} & \textit{0.184} \\
            &  &  &  & GIV & 0.113 & 0.537 \\
        \midrule
        \multirow{5}{*}{\makecell[l]{Linear Disjoint (no $U \to X$)}} 
            & 0.267 & 0.207 & 0.327 & TrueIV & \textbf{0.148} & \textit{0.280} \\
            &  &  &  & ZNet & 0.222 & 0.469 \\
            &  &  &  & AutoIV & 0.199 & 0.775 \\
            &  &  &  & VIV & 0.190 & \textbf{0.269} \\
            &  &  &  & GIV & \textit{0.168} & 0.521 \\
        \midrule
        \multirow{5}{*}{\makecell[l]{Linear Latent Categorical}} 
            & 0.160 & 0.157 & 1.118 & TrueIV & 0.101 & \textit{0.171} \\
            &  &  &  & ZNet & \textbf{0.090} & \textbf{0.080} \\
            &  &  &  & AutoIV & 0.134 & 0.626 \\
            &  &  &  & VIV & \textit{0.092} & 0.353 \\
            &  &  &  & GIV & 0.095 & 0.492 \\
        \midrule
        \multirow{5}{*}{\makecell[l]{Linear Latent Categorical (no $U \to X$)}} 
            & 0.390 & 0.372 & 1.088 & TrueIV & 0.389 & \textbf{0.384} \\
            &  &  &  & ZNet & 0.324 & 0.453 \\
            &  &  &  & AutoIV & 0.343 & 0.723 \\
            &  &  &  & VIV & \textit{0.309} & \textit{0.411} \\
            &  &  &  & GIV & \textbf{0.298} & 0.535 \\
        \midrule
        \multirow{5}{*}{\makecell[l]{Linear Mixed}} 
            & 0.790 & 0.570 & 0.448 & TrueIV & \textit{0.660} & \textbf{0.592} \\
            &  &  &  & ZNet & \textbf{0.537} & 0.844 \\
            &  &  &  & AutoIV & 0.690 & 0.886 \\
            &  &  &  & VIV & 0.716 & 0.859 \\
            &  &  &  & GIV & 0.685 & \textit{0.750} \\
        \midrule
        \multirow{5}{*}{\makecell[l]{Linear Mixed (no $U \to X$)}} 
            & 0.390 & 0.298 & 0.326 & TrueIV & \textbf{0.314} & 0.446 \\
            &  &  &  & ZNet & \textit{0.314} & 0.535 \\
            &  &  &  & AutoIV & 0.344 & 0.463 \\
            &  &  &  & VIV & 0.375 & \textit{0.411} \\
            &  &  &  & GIV & 0.370 & \textbf{0.324} \\
        \midrule
        \multirow{5}{*}{\makecell[l]{Linear No Candidate}} 
            & 1.749 & 1.375 & -- & TrueIV & -- & -- \\
            &  &  &  & ZNet & \textbf{1.306} & 1.724 \\
            &  &  &  & AutoIV & \textit{1.418} & 1.630 \\
            &  &  &  & VIV & 1.460 & \textbf{1.429} \\
            &  &  &  & GIV & 1.491 & \textit{1.546} \\
        \midrule
        \multirow{5}{*}{\makecell[l]{Linear No Candidate (no $U \to X$)}} 
            & 0.291 & 0.365 & -- & TrueIV & -- & -- \\
            &  &  &  & ZNet & \textbf{0.361} & \textit{0.713} \\
            &  &  &  & AutoIV & 0.405 & 0.831 \\
            &  &  &  & VIV & \textit{0.367} & \textbf{0.499} \\
            &  &  &  & GIV & 0.435 & 1.193 \\
        \midrule
        \multirow{5}{*}{\makecell[l]{Linear No Candidate (no $U$)}} 
            & 0.060 & 0.092 & -- & TrueIV & -- & -- \\
            &  &  &  & ZNet & \textbf{0.037} & \textit{0.245} \\
            &  &  &  & AutoIV & 0.064 & 0.823 \\
            &  &  &  & VIV & \textit{0.043} & \textbf{0.190} \\
            &  &  &  & GIV & 0.052 & 0.556 \\
        \bottomrule
    \end{tabular}
    \label{tab:appendix-ate-linear-results}
\end{table}
\begin{table}[h!]
\footnotesize
    \centering
    \caption{ATE Results on Synthetic Non-Linear Datasets}
    \begin{tabular}{lccclcc}
        \toprule
        \textbf{Dataset} & \textbf{Diff Means} & \textbf{TARNet} & \textbf{TSLS} & \textbf{IV Method} & \textbf{DeepIV} & \textbf{DF IV} \\
        \midrule
        \multirow{5}{*}{\makecell[l]{Non-linear Disjoint}} 
            & 0.845 & 0.571 & 0.424 & TrueIV & 0.586 & \textit{0.602} \\
            &  &  &  & ZNet & \textit{0.553} & 0.706 \\
            &  &  &  & AutoIV & 0.656 & 0.869 \\
            &  &  &  & VIV & 0.633 & \textbf{0.558} \\
            &  &  &  & GIV & \textbf{0.551} & 0.854 \\
        \midrule
        \multirow{5}{*}{\makecell[l]{Non-linear Disjoint (no $U \to X$)}} 
            & 0.172 & 0.255 & 0.349 & TrueIV & \textbf{0.202} & 0.661 \\
            &  &  &  & ZNet & 0.274 & \textit{0.354} \\
            &  &  &  & AutoIV & 0.283 & 0.606 \\
            &  &  &  & VIV & 0.218 & \textbf{0.339} \\
            &  &  &  & GIV & \textit{0.215} & 1.951 \\
        \midrule
        \multirow{5}{*}{\makecell[l]{Non-linear Latent Categorical}} 
            & 1.441 & 0.784 & 12.998 & TrueIV & 0.697 & 1.758 \\
            &  &  &  & ZNet & 0.689 & 0.784 \\
            &  &  &  & AutoIV & \textbf{0.581} & 1.080 \\
            &  &  &  & VIV & 0.798 & \textbf{0.678} \\
            &  &  &  & GIV & \textit{0.647} & \textit{0.727} \\
        \midrule
        \multirow{5}{*}{\makecell[l]{Non-linear Latent Categorical (no $U \to X$)}} 
            & 0.431 & 0.405 & 1.569 & TrueIV & 0.375 & 0.499 \\
            &  &  &  & ZNet & 0.350 & \textit{0.283} \\
            &  &  &  & AutoIV & 0.376 & 0.319 \\
            &  &  &  & VIV & \textit{0.331} & \textbf{0.253} \\
            &  &  &  & GIV & \textbf{0.299} & 0.810 \\
        \midrule
        \multirow{5}{*}{\makecell[l]{Non-linear Mixed}} 
            & 1.501 & 1.190 & 1.570 & TrueIV & \textbf{1.051} & 1.118 \\
            &  &  &  & ZNet & 1.200 & 1.005 \\
            &  &  &  & AutoIV & 1.156 & \textit{0.971} \\
            &  &  &  & VIV & 1.179 & 1.397 \\
            &  &  &  & GIV & \textit{1.098} & \textbf{0.938} \\
        \midrule
        \multirow{5}{*}{\makecell[l]{Non-linear Mixed (no $U \to X$)}} 
            & 0.205 & 0.342 & 0.535 & TrueIV & \textbf{0.232} & 0.564 \\
            &  &  &  & ZNet & 0.379 & \textbf{0.327} \\
            &  &  &  & AutoIV & 0.250 & \textit{0.351} \\
            &  &  &  & VIV & \textit{0.245} & 1.786 \\
            &  &  &  & GIV & 0.254 & 0.635 \\
        \midrule
        \multirow{5}{*}{\makecell[l]{Non-linear No Candidate}} 
            & 0.658 & 0.542 & -- & TrueIV & -- & -- \\
            &  &  &  & ZNet & \textbf{0.478} & \textbf{0.457} \\
            &  &  &  & AutoIV & 0.690 & \textit{0.567} \\
            &  &  &  & VIV & \textit{0.654} & 1.788 \\
            &  &  &  & GIV & 0.698 & 1.040 \\
        \midrule
        \multirow{5}{*}{\makecell[l]{Non-linear No Candidate (no $U \to X$)}} 
            & 0.216 & 0.193 & -- & TrueIV & -- & -- \\
            &  &  &  & ZNet & \textbf{0.121} & \textbf{0.200} \\
            &  &  &  & AutoIV & 0.155 & 0.321 \\
            &  &  &  & VIV & \textit{0.133} & \textit{0.229} \\
            &  &  &  & GIV & 0.190 & 0.528 \\
        \midrule
        \multirow{5}{*}{\makecell[l]{Non-linear No Candidate (no $U$)}} 
            & 0.109 & 0.105 & -- & TrueIV & -- & -- \\
            &  &  &  & ZNet & \textit{0.064} & \textit{0.387} \\
            &  &  &  & AutoIV & 0.079 & 0.513 \\
            &  &  &  & VIV & \textbf{0.046} & \textbf{0.110} \\
            &  &  &  & GIV & 0.068 & 0.433 \\
        \bottomrule
    \end{tabular} 
    \label{tab:appendix-ate-nonlinear-results}
\end{table}

\begin{table}[h!]
\footnotesize
    \caption{PEHE on Linear Synthetic Dataset.}
    \centering
    \begin{tabular}{lclcc}
        \toprule
        \textbf{Dataset} & \textbf{TARNet} & \textbf{IV Method} & \textbf{DeepIV} & \textbf{DF IV} \\
        \midrule
        \multirow{5}{*}{\makecell[l]{Linear Disjoint}} 
            & 0.142 & TrueIV & 0.165 & \textbf{0.187} \\
            &  & ZNet & 0.189 & 0.240 \\
            &  & AutoIV & \textbf{0.150} & 0.323 \\
            &  & VIV & \textit{0.153} & \textit{0.209} \\
            &  & GIV & 0.188 & 0.445 \\
        \midrule
        \multirow{5}{*}{\makecell[l]{Linear Disjoint (no $U \to X$)}} 
            & 0.176 & TrueIV & \textbf{0.202} & \textbf{0.272} \\
            &  & ZNet & \textit{0.203} & 0.431 \\
            &  & AutoIV & 0.224 & 0.541 \\
            &  & VIV & 0.215 & \textit{0.273} \\
            &  & GIV & 0.237 & 0.428 \\
        \midrule
        \multirow{5}{*}{\makecell[l]{Linear Latent Categorical}} 
            & 0.213 & TrueIV & 0.264 & \textit{0.259} \\
            &  & ZNet & \textit{0.239} & \textbf{0.212} \\
            &  & AutoIV & 0.283 & 0.692 \\
            &  & VIV & 0.263 & 0.470 \\
            &  & GIV & \textbf{0.230} & 0.627 \\
        \midrule
        \multirow{5}{*}{\makecell[l]{Linear Latent Categorical (no $U \to X$)}} 
            & 0.421 & TrueIV & 0.524 & \textbf{0.409} \\
            &  & ZNet & \textit{0.477} & 0.502 \\
            &  & AutoIV & 0.508 & 0.676 \\
            &  & VIV & 0.495 & \textit{0.451} \\
            &  & GIV & \textbf{0.471} & 0.501 \\
        \midrule
        \multirow{5}{*}{\makecell[l]{Linear Mixed}} 
            & 0.681 & TrueIV & \textit{0.805} & \textbf{0.800} \\
            &  & ZNet & \textbf{0.692} & 1.167 \\
            &  & AutoIV & 0.828 & 0.943 \\
            &  & VIV & 0.836 & 0.969 \\
            &  & GIV & 0.824 & \textit{0.855} \\
        \midrule
        \multirow{5}{*}{\makecell[l]{Linear Mixed (no $U \to X$)}} 
            & 0.385 & TrueIV & \textbf{0.527} & 0.672 \\
            &  & ZNet & 0.593 & 0.817 \\
            &  & AutoIV & 0.566 & 0.658 \\
            &  & VIV & 0.599 & \textit{0.536} \\
            &  & GIV & \textit{0.540} & \textbf{0.493} \\
        \midrule
        \multirow{5}{*}{\makecell[l]{Linear No Candidate}} 
            & 0.726 & TrueIV & -- & -- \\
            &  & ZNet & \textbf{0.741} & 0.935 \\
            &  & AutoIV & \textit{0.799} & \textit{0.793} \\
            &  & VIV & 0.825 & \textbf{0.777} \\
            &  & GIV & 0.799 & 0.801 \\
        \midrule
        \multirow{5}{*}{\makecell[l]{Linear No Candidate (no $U \to X$)}} 
            & 0.428 & TrueIV & -- & -- \\
            &  & ZNet & \textbf{0.464} & 0.912 \\
            &  & AutoIV & 0.563 & \textit{0.884} \\
            &  & VIV & \textit{0.530} & \textbf{0.587} \\
            &  & GIV & 0.568 & 1.321 \\
        \midrule
        \multirow{5}{*}{\makecell[l]{Linear No Candidate (no $U$)}} 
            & 0.178 & TrueIV & -- & -- \\
            &  & ZNet & \textit{0.229} & \textit{0.570} \\
            &  & AutoIV & 0.342 & 1.553 \\
            &  & VIV & \textbf{0.224} & \textbf{0.452} \\
            &  & GIV & 0.313 & 0.932 \\
        \bottomrule
    \end{tabular} 
    \label{tab:appendix-pehe-linear}
\end{table}
\begin{table}[h!]
\footnotesize
    \caption{PEHE on Non-linear Synthetic Dataset.}
    \centering
    \begin{tabular}{lclcc}
        \toprule
        \textbf{Dataset} & \textbf{TARNet} & \textbf{IV Method} & \textbf{DeepIV} & \textbf{DF IV} \\
        \midrule
        \multirow{5}{*}{\makecell[l]{Non-linear Disjoint}} 
            & 0.788 & TrueIV & 0.762 & 0.942 \\
            &  & ZNet & \textbf{0.733} & \textit{0.855} \\
            &  & AutoIV & 0.805 & 1.028 \\
            &  & VIV & 0.818 & \textbf{0.700} \\
            &  & GIV & \textit{0.753} & 0.930 \\
        \midrule
        \multirow{5}{*}{\makecell[l]{Non-linear Disjoint (no $U \to X$)}} 
            & 0.548 & TrueIV & \textbf{0.509} & \textit{0.834} \\
            &  & ZNet & 0.528 & 1.253 \\
            &  & AutoIV & 0.556 & 0.846 \\
            &  & VIV & 0.523 & \textbf{0.567} \\
            &  & GIV & \textit{0.509} & 2.245 \\
        \midrule
        \multirow{5}{*}{\makecell[l]{Non-linear Latent Categorical}} 
            & 0.529 & TrueIV & 0.561 & 1.121 \\
            &  & ZNet & \textbf{0.515} & \textit{0.548} \\
            &  & AutoIV & \textit{0.519} & 0.954 \\
            &  & VIV & 0.626 & 0.552 \\
            &  & GIV & 0.524 & \textbf{0.520} \\
        \midrule
        \multirow{5}{*}{\makecell[l]{Non-linear Latent Categorical (no $U \to X$)}} 
            & 0.430 & TrueIV & 0.493 & 0.472 \\
            &  & ZNet & \textbf{0.436} & \textit{0.304} \\
            &  & AutoIV & 0.527 & 0.397 \\
            &  & VIV & 0.455 & \textbf{0.303} \\
            &  & GIV & \textit{0.446} & 0.709 \\
        \midrule
        \multirow{5}{*}{\makecell[l]{Non-linear Mixed}} 
            & 0.965 & TrueIV & \textbf{0.818} & 0.786 \\
            &  & ZNet & 0.988 & 0.866 \\
            &  & AutoIV & 0.956 & \textit{0.777} \\
            &  & VIV & 1.005 & 0.882 \\
            &  & GIV & \textit{0.887} & \textbf{0.704} \\
        \midrule
        \multirow{5}{*}{\makecell[l]{Non-linear Mixed (no $U \to X$)}} 
            & 0.952 & TrueIV & 0.842 & 0.963 \\
            &  & ZNet & 0.854 & \textit{0.859} \\
            &  & AutoIV & \textit{0.841} & \textbf{0.846} \\
            &  & VIV & 0.858 & 2.192 \\
            &  & GIV & \textbf{0.820} & 1.245 \\
        \midrule
        \multirow{5}{*}{\makecell[l]{Non-linear No Candidate}} 
            & 0.648 & TrueIV & -- & -- \\
            &  & ZNet & \textbf{0.624} & \textbf{0.615} \\
            &  & AutoIV & 0.730 & \textit{0.622} \\
            &  & VIV & 0.724 & 1.417 \\
            &  & GIV & \textit{0.709} & 0.822 \\
        \midrule
        \multirow{5}{*}{\makecell[l]{Non-linear No Candidate (no $U \to X$)}} 
            & 0.833 & TrueIV & -- & -- \\
            &  & ZNet & 0.818 & 0.802 \\
            &  & AutoIV & 0.820 & \textit{0.801} \\
            &  & VIV & \textit{0.805} & \textbf{0.775} \\
            &  & GIV & \textbf{0.793} & 1.249 \\
        \midrule
        \multirow{5}{*}{\makecell[l]{Non-linear No Candidate (no $U$)}} 
            & 1.165 & TrueIV & -- & -- \\
            &  & ZNet & \textbf{1.081} & 1.393 \\
            &  & AutoIV & 1.158 & 1.415 \\
            &  & VIV & 1.181 & \textbf{1.070} \\
            &  & GIV & \textit{1.144} & \textit{1.146} \\
        \bottomrule
    \end{tabular} 
    \label{tab:appendix-pehe-nonlinear}
\end{table}

\begin{table}[h!]
\footnotesize
    \tiny
    \caption{Instrument Strength and Validity Linear Synthetic Dataset.}
    \centering
    \begin{tabular}{llcccc}
        \toprule
        \textbf{Dataset} & \textbf{IV Method} & \makecell{\textbf{F-Stat(Z,T)} \\ \textbf{(Relevance)} \\ (Train/Val/Test)} & \makecell{\textbf{Corr(Z,C)} \\ \textbf{(Independence)} \\ (Train/Val/Test)} & \makecell{\textbf{Corr(Z,Y-Yhat)} \\ \textbf{(Exogeneity)} \\ (Train/Val/Test)} & \makecell{\textbf{Corr(Z,U)} \\ \textbf{(Independence)} \\ (Train/Val/Test)} \\
        \midrule
        \multirow{5}{*}{\makecell[l]{Linear Disjoint}} 
            & TrueIV & 9.792 / 3.215 / 3.657 & 0.027 / 0.051 / 0.124 & 0.032 / 0.069 / 0.095 & 0.030 / 0.062 / 0.045 \\
            & ZNet & 28.439 / 2.444 / 3.085 & 0.128 / 0.141 / 0.180 & 0.033 / 0.068 / 0.090 & 0.158 / 0.166 / 0.170 \\
            & AutoIV & 5.275 / 1.880 / 2.731 & 0.208 / 0.221 / 0.243 & 0.000 / 0.000 / 0.000 & 0.104 / 0.109 / 0.098 \\
            & VIV & 17.719 / 5.397 / 4.568 & 0.032 / 0.057 / 0.112 & 0.032 / 0.049 / 0.101 & 0.036 / 0.050 / 0.075 \\
            & GIV & 8.737 / 2.217 / 3.190 & 0.139 / 0.143 / 0.173 & 0.032 / 0.039 / 0.067 & 0.080 / 0.084 / 0.111 \\
        \midrule
        \multirow{5}{*}{\makecell[l]{Linear Disjoint (no $U \to X$)}} 
            & TrueIV & 17.167 / 7.477 / 6.471 & 0.027 / 0.051 / 0.124 & 0.051 / 0.056 / 0.116 & 0.035 / 0.049 / 0.095 \\
            & ZNet & 45.214 / 8.188 / 3.851 & 0.080 / 0.111 / 0.149 & 0.025 / 0.052 / 0.095 & 0.029 / 0.056 / 0.093 \\
            & AutoIV & 9.318 / 2.783 / 2.159 & 0.212 / 0.222 / 0.255 & 0.000 / 0.000 / 0.000 & 0.030 / 0.062 / 0.091 \\
            & VIV & 16.739 / 3.926 / 3.641 & 0.030 / 0.058 / 0.119 & 0.025 / 0.065 / 0.109 & 0.036 / 0.053 / 0.080 \\
            & GIV & 0.913 / 1.251 / 2.414 & 0.148 / 0.149 / 0.190 & 0.024 / 0.045 / 0.072 & 0.039 / 0.062 / 0.087 \\
        \midrule
        \multirow{5}{*}{\makecell[l]{Linear Latent Categorical}} 
            & TrueIV & 165.204 / 53.555 / 19.690 & 0.202 / 0.206 / 0.230 & 0.011 / 0.014 / 0.028 & 0.035 / 0.048 / 0.076 \\
            & ZNet & 179.695 / 25.554 / 9.951 & 0.134 / 0.134 / 0.165 & 0.014 / 0.039 / 0.071 & 0.074 / 0.081 / 0.098 \\
            & AutoIV & 68.529 / 18.436 / 6.537 & 0.258 / 0.247 / 0.263 & 0.000 / 0.000 / 0.000 & 0.144 / 0.154 / 0.156 \\
            & VIV & 17.504 / 5.266 / 5.422 & 0.035 / 0.064 / 0.118 & 0.030 / 0.039 / 0.097 & 0.036 / 0.056 / 0.082 \\
            & GIV & 97.670 / 25.682 / 12.462 & 0.140 / 0.157 / 0.178 & 0.020 / 0.024 / 0.065 & 0.036 / 0.067 / 0.081 \\
        \midrule
        \multirow{5}{*}{\makecell[l]{Linear Latent Categorical (no $U \to X$)}} 
            & TrueIV & 199.602 / 65.321 / 24.395 & 0.204 / 0.212 / 0.235 & 0.006 / 0.016 / 0.031 & 0.036 / 0.049 / 0.065 \\
            & ZNet & 300.974 / 25.263 / 11.940 & 0.111 / 0.117 / 0.138 & 0.010 / 0.053 / 0.085 & 0.043 / 0.058 / 0.073 \\
            & AutoIV & 91.941 / 28.877 / 8.274 & 0.249 / 0.263 / 0.266 & 0.000 / 0.000 / 0.000 & 0.031 / 0.054 / 0.074 \\
            & VIV & 16.884 / 4.987 / 4.814 & 0.034 / 0.058 / 0.115 & 0.024 / 0.042 / 0.099 & 0.031 / 0.064 / 0.081 \\
            & GIV & 35.204 / 24.341 / 9.507 & 0.146 / 0.146 / 0.191 & 0.017 / 0.038 / 0.056 & 0.029 / 0.054 / 0.082 \\
        \midrule
        \multirow{5}{*}{\makecell[l]{Linear Mixed}} 
            & TrueIV & 8.070 / 2.816 / 3.181 & 0.027 / 0.051 / 0.124 & 0.081 / 0.099 / 0.113 & 0.030 / 0.062 / 0.045 \\
            & ZNet & 221.401 / 25.663 / 20.673 & 0.209 / 0.208 / 0.238 & 0.013 / 0.053 / 0.084 & 0.284 / 0.240 / 0.303 \\
            & AutoIV & 103.603 / 22.816 / 23.096 & 0.209 / 0.211 / 0.235 & 0.000 / 0.000 / 0.000 & 0.253 / 0.219 / 0.289 \\
            & VIV & 18.179 / 5.523 / 4.300 & 0.033 / 0.053 / 0.111 & 0.030 / 0.044 / 0.112 & 0.050 / 0.072 / 0.091 \\
            & GIV & 43.053 / 8.902 / 8.220 & 0.147 / 0.150 / 0.195 & 0.026 / 0.041 / 0.080 & 0.124 / 0.117 / 0.157 \\
        \midrule
        \multirow{5}{*}{\makecell[l]{Linear Mixed (no $U \to X$)}} 
            & TrueIV & 12.066 / 5.101 / 3.489 & 0.027 / 0.051 / 0.124 & 0.063 / 0.081 / 0.097 & 0.027 / 0.060 / 0.079 \\
            & ZNet & 133.538 / 16.742 / 8.084 & 0.150 / 0.155 / 0.186 & 0.025 / 0.077 / 0.114 & 0.045 / 0.055 / 0.094 \\
            & AutoIV & 104.522 / 31.072 / 17.727 & 0.237 / 0.241 / 0.261 & 0.000 / 0.000 / 0.000 & 0.029 / 0.049 / 0.094 \\
            & VIV & 16.356 / 7.037 / 3.971 & 0.033 / 0.060 / 0.117 & 0.028 / 0.053 / 0.101 & 0.042 / 0.055 / 0.086 \\
            & GIV & 12.316 / 2.874 / 4.237 & 0.150 / 0.159 / 0.194 & 0.027 / 0.026 / 0.063 & 0.029 / 0.042 / 0.095 \\
        \midrule
        \multirow{5}{*}{\makecell[l]{Linear No Candidate}} 
            & TrueIV & -- & -- & -- & -- \\
            & ZNet & 141.454 / 17.320 / 8.324 & 0.150 / 0.168 / 0.199 & 0.019 / 0.070 / 0.079 & 0.327 / 0.310 / 0.295 \\
            & AutoIV & 88.696 / 23.879 / 11.640 & 0.245 / 0.240 / 0.272 & 0.000 / 0.000 / 0.000 & 0.404 / 0.399 / 0.373 \\
            & VIV & 18.248 / 4.797 / 4.426 & 0.033 / 0.059 / 0.115 & 0.039 / 0.062 / 0.098 & 0.057 / 0.074 / 0.098 \\
            & GIV & 52.157 / 15.182 / 5.950 & 0.145 / 0.141 / 0.186 & 0.029 / 0.039 / 0.075 & 0.124 / 0.144 / 0.134 \\
        \midrule
        \multirow{5}{*}{\makecell[l]{Linear No Candidate (no $U \to X$)}} 
            & TrueIV & -- & -- & -- & -- \\
            & ZNet & 71.331 / 2.770 / 3.064 & 0.166 / 0.181 / 0.213 & 0.024 / 0.069 / 0.095 & 0.044 / 0.060 / 0.079 \\
            & AutoIV & 16.775 / 2.015 / 2.158 & 0.236 / 0.242 / 0.254 & 0.000 / 0.000 / 0.000 & 0.041 / 0.057 / 0.085 \\
            & VIV & 18.499 / 7.769 / 4.560 & 0.032 / 0.065 / 0.115 & 0.037 / 0.052 / 0.099 & 0.037 / 0.068 / 0.072 \\
            & GIV & 0.850 / 0.970 / 1.879 & 0.154 / 0.166 / 0.207 & 0.020 / 0.033 / 0.066 & 0.028 / 0.058 / 0.084 \\
        \midrule
        \multirow{5}{*}{\makecell[l]{Linear No Candidate (no $U$)}} 
            & TrueIV & -- & -- & -- & -- \\
            & ZNet & 115.223 / 2.995 / 2.728 & 0.147 / 0.156 / 0.190 & 0.031 / 0.062 / 0.082 & -- / -- / -- \\
            & AutoIV & 12.631 / 2.437 / 2.154 & 0.216 / 0.232 / 0.250 & 0.000 / 0.000 / 0.000 & -- / -- / -- \\
            & VIV & 21.595 / 5.473 / 4.768 & 0.031 / 0.054 / 0.116 & 0.037 / 0.047 / 0.113 & -- / -- / -- \\
            & GIV & 1.574 / 0.649 / 1.787 & 0.161 / 0.164 / 0.200 & 0.022 / 0.018 / 0.062 & -- / -- / -- \\
        \bottomrule
    \end{tabular}
    \label{tab:appendix-linear-instrument-strength}
\end{table}
\begin{table}[h!]
    \tiny
    \caption{Instrument Strength and Validity on Non-linear Synthetic Dataset.}
    \centering
    \begin{tabular}{llcccc}
        \toprule
        \textbf{Dataset} & \textbf{IV Method} & \makecell{\textbf{F-Stat(Z,T)} \\ \textbf{(Relevance)} \\ (Train/Val/Test)} & \makecell{\textbf{Corr(Z,C)} \\ \textbf{(Independence)} \\ (Train/Val/Test)} & \makecell{\textbf{Corr(Z,Y-Yhat)} \\ \textbf{(Exogeneity)} \\ (Train/Val/Test)} & \makecell{\textbf{Corr(Z,U)} \\ \textbf{(Independence)} \\ (Train/Val/Test)} \\
        \midrule
        \multirow{5}{*}{\makecell[l]{Non-linear Disjoint}} 
            & TrueIV & 10.436 / 4.195 / 2.690 & 0.027 / 0.051 / 0.124 & 0.079 / 0.078 / 0.108 & 0.030 / 0.062 / 0.045 \\
            & ZNet & 298.079 / 24.161 / 14.563 & 0.163 / 0.178 / 0.192 & 0.031 / 0.049 / 0.080 & 0.285 / 0.256 / 0.257 \\
            & AutoIV & 102.265 / 24.380 / 18.903 & 0.206 / 0.217 / 0.227 & 0.000 / 0.000 / 0.000 & 0.278 / 0.241 / 0.295 \\
            & VIV & 16.532 / 5.867 / 5.673 & 0.033 / 0.058 / 0.114 & 0.034 / 0.047 / 0.094 & 0.059 / 0.069 / 0.102 \\
            & GIV & 17.338 / 5.966 / 5.895 & 0.141 / 0.150 / 0.181 & 0.036 / 0.048 / 0.074 & 0.144 / 0.140 / 0.176 \\
        \midrule
        \multirow{5}{*}{\makecell[l]{Non-linear Disjoint (no $U \to X$)}} 
            & TrueIV & 19.393 / 6.986 / 5.568 & 0.027 / 0.051 / 0.124 & 0.028 / 0.055 / 0.115 & 0.030 / 0.044 / 0.094 \\
            & ZNet & 70.675 / 7.622 / 4.336 & 0.143 / 0.157 / 0.174 & 0.017 / 0.053 / 0.105 & 0.033 / 0.046 / 0.081 \\
            & AutoIV & 40.857 / 6.378 / 3.899 & 0.236 / 0.236 / 0.260 & 0.000 / 0.000 / 0.000 & 0.031 / 0.043 / 0.090 \\
            & VIV & 17.436 / 5.063 / 4.857 & 0.034 / 0.056 / 0.117 & 0.042 / 0.051 / 0.101 & 0.034 / 0.058 / 0.076 \\
            & GIV & 2.574 / 1.176 / 1.936 & 0.140 / 0.148 / 0.187 & 0.042 / 0.039 / 0.067 & 0.035 / 0.060 / 0.072 \\
        \midrule
        \multirow{5}{*}{\makecell[l]{Non-linear Latent Categorical}} 
            & TrueIV & 96.409 / 39.441 / 8.684 & 0.204 / 0.212 / 0.235 & 0.009 / 0.011 / 0.025 & 0.064 / 0.083 / 0.022 \\
            & ZNet & 120.953 / 20.441 / 7.507 & 0.173 / 0.172 / 0.188 & 0.030 / 0.069 / 0.099 & 0.174 / 0.178 / 0.179 \\
            & AutoIV & 58.730 / 12.420 / 9.422 & 0.213 / 0.221 / 0.236 & 0.000 / 0.000 / 0.000 & 0.251 / 0.231 / 0.250 \\
            & VIV & 15.337 / 3.719 / 4.762 & 0.034 / 0.061 / 0.113 & 0.034 / 0.043 / 0.091 & 0.037 / 0.063 / 0.083 \\
            & GIV & 57.630 / 20.981 / 10.098 & 0.131 / 0.142 / 0.169 & 0.025 / 0.035 / 0.068 & 0.118 / 0.103 / 0.149 \\
        \midrule
        \multirow{5}{*}{\makecell[l]{Non-linear Latent Categorical (no $U \to X$)}} 
            & TrueIV & 129.011 / 37.750 / 14.270 & 0.204 / 0.212 / 0.235 & 0.005 / 0.020 / 0.029 & 0.033 / 0.050 / 0.071 \\
            & ZNet & 120.243 / 19.860 / 9.497 & 0.148 / 0.150 / 0.193 & 0.019 / 0.054 / 0.077 & 0.025 / 0.054 / 0.071 \\
            & AutoIV & 79.648 / 20.792 / 7.362 & 0.253 / 0.267 / 0.273 & 0.000 / 0.000 / 0.000 & 0.032 / 0.049 / 0.091 \\
            & VIV & 16.644 / 6.861 / 4.575 & 0.033 / 0.061 / 0.117 & 0.033 / 0.057 / 0.106 & 0.040 / 0.065 / 0.087 \\
            & GIV & 34.207 / 11.849 / 5.966 & 0.141 / 0.146 / 0.169 & 0.017 / 0.039 / 0.058 & 0.028 / 0.057 / 0.087 \\
        \midrule
        \multirow{5}{*}{\makecell[l]{Non-linear Mixed}} 
            & TrueIV & 4.313 / 1.787 / 2.142 & 0.027 / 0.051 / 0.124 & 0.058 / 0.098 / 0.102 & 0.030 / 0.062 / 0.045 \\
            & ZNet & 144.440 / 18.633 / 15.304 & 0.167 / 0.156 / 0.189 & 0.036 / 0.083 / 0.107 & 0.201 / 0.185 / 0.221 \\
            & AutoIV & 94.883 / 17.895 / 17.240 & 0.210 / 0.222 / 0.238 & 0.000 / 0.000 / 0.000 & 0.224 / 0.210 / 0.248 \\
            & VIV & 15.459 / 4.695 / 3.970 & 0.035 / 0.058 / 0.111 & 0.043 / 0.068 / 0.117 & 0.052 / 0.062 / 0.091 \\
            & GIV & 56.539 / 18.722 / 9.017 & 0.138 / 0.145 / 0.194 & 0.034 / 0.043 / 0.077 & 0.136 / 0.126 / 0.153 \\
        \midrule
        \multirow{5}{*}{\makecell[l]{Non-linear Mixed (no $U \to X$)}} 
            & TrueIV & 16.903 / 4.576 / 3.954 & 0.027 / 0.051 / 0.124 & 0.026 / 0.057 / 0.120 & 0.027 / 0.055 / 0.085 \\
            & ZNet & 281.071 / 28.446 / 19.673 & 0.236 / 0.239 / 0.275 & 0.020 / 0.038 / 0.071 & 0.043 / 0.059 / 0.067 \\
            & AutoIV & 163.159 / 34.251 / 23.897 & 0.229 / 0.235 / 0.257 & 0.000 / 0.000 / 0.000 & 0.031 / 0.063 / 0.069 \\
            & VIV & 19.235 / 3.738 / 4.668 & 0.035 / 0.059 / 0.114 & 0.034 / 0.065 / 0.099 & 0.032 / 0.064 / 0.087 \\
            & GIV & 24.444 / 8.313 / 5.430 & 0.137 / 0.160 / 0.188 & 0.024 / 0.036 / 0.061 & 0.032 / 0.053 / 0.079 \\
        \midrule
        \multirow{5}{*}{\makecell[l]{Non-linear No Candidate}} 
            & TrueIV & -- & -- & -- & -- \\
            & ZNet & 54.003 / 9.565 / 4.068 & 0.144 / 0.186 / 0.187 & 0.022 / 0.061 / 0.091 & 0.198 / 0.195 / 0.191 \\
            & AutoIV & 62.921 / 16.672 / 6.355 & 0.202 / 0.201 / 0.236 & 0.000 / 0.000 / 0.000 & 0.322 / 0.324 / 0.318 \\
            & VIV & 20.382 / 4.410 / 4.932 & 0.034 / 0.057 / 0.115 & 0.031 / 0.047 / 0.096 & 0.040 / 0.060 / 0.090 \\
            & GIV & 23.479 / 3.748 / 3.176 & 0.145 / 0.147 / 0.182 & 0.029 / 0.028 / 0.069 & 0.091 / 0.098 / 0.099 \\
        \midrule
        \multirow{5}{*}{\makecell[l]{Non-linear No Candidate (no $U \to X$)}} 
            & TrueIV & -- & -- & -- & -- \\
            & ZNet & 100.546 / 7.451 / 6.169 & 0.114 / 0.121 / 0.171 & 0.033 / 0.063 / 0.081 & 0.043 / 0.052 / 0.075 \\
            & AutoIV & 55.722 / 7.239 / 7.235 & 0.215 / 0.217 / 0.246 & 0.000 / 0.000 / 0.000 & 0.033 / 0.054 / 0.073 \\
            & VIV & 19.903 / 7.638 / 5.995 & 0.033 / 0.062 / 0.118 & 0.039 / 0.049 / 0.108 & 0.039 / 0.060 / 0.073 \\
            & GIV & 71.468 / 20.206 / 8.957 & 0.151 / 0.153 / 0.190 & 0.021 / 0.036 / 0.066 & 0.041 / 0.058 / 0.086 \\
        \midrule
        \multirow{5}{*}{\makecell[l]{Non-linear No Candidate (no $U$)}} 
            & TrueIV & -- & -- & -- & -- \\
            & ZNet & 97.217 / 3.587 / 3.745 & 0.144 / 0.149 / 0.207 & 0.032 / 0.039 / 0.110 & -- / -- / -- \\
            & AutoIV & 40.972 / 4.580 / 5.317 & 0.215 / 0.222 / 0.251 & 0.000 / 0.000 / 0.000 & -- / -- / -- \\
            & VIV & 18.745 / 3.922 / 5.198 & 0.031 / 0.063 / 0.109 & 0.022 / 0.063 / 0.118 & -- / -- / -- \\
            & GIV & 87.656 / 21.257 / 17.612 & 0.145 / 0.151 / 0.184 & 0.031 / 0.036 / 0.080 & -- / -- / -- \\
        \bottomrule
    \end{tabular}
    \label{tab:appendix-nonlinear-instrument-strength}
\end{table}

\begin{figure}[h!]
\footnotesize
    \centering
    \includegraphics[width=.8\linewidth]{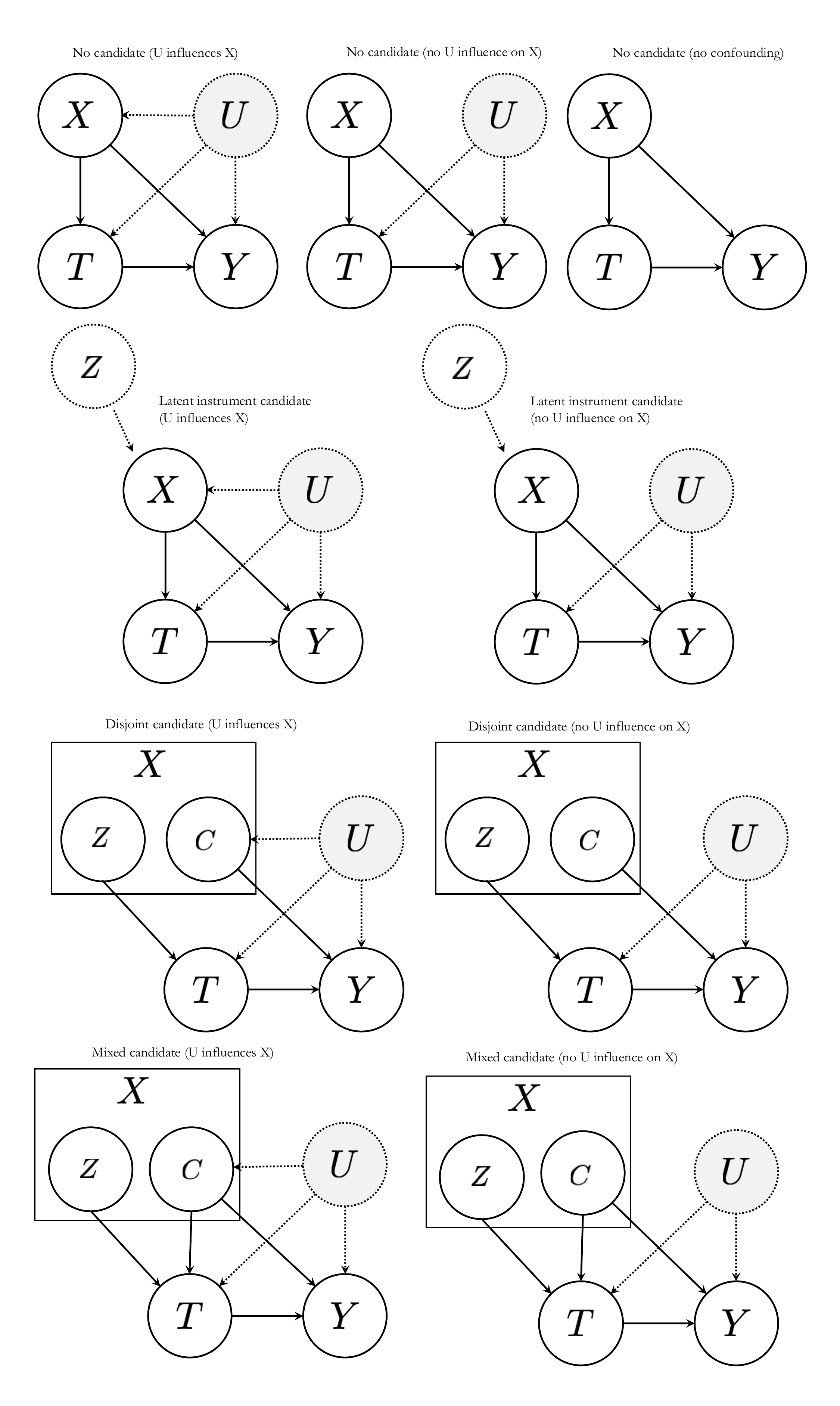}
    \caption{{\textbf{Directed acyclic graphs (DAGs) demonstrating the various data generation processes on which ZNet is evaluated}. Linear and non-linear relationships are constructed for each DAG giving 18 total datasets for evaluation. Maintext results focus on cases where $U$ influences $X$ as this is more challenging, more general, and unique to ZNet.}}
    \label{fig:dags}
\end{figure}

\end{document}